\documentclass[journal,10pt]{IEEEtran}
\usepackage{lineno}
\modulolinenumbers[5]
\usepackage{amsfonts}
\usepackage{amsmath}
\usepackage{bm}
\usepackage{url}
\usepackage{mathrsfs}
\usepackage{mathtools}
\usepackage{algorithm}
\usepackage{algorithmicx}
\usepackage{pseudocode}
\usepackage{algpseudocode}
\usepackage{multirow}
\usepackage{bbm}
\usepackage{graphicx}
\usepackage{enumitem}
\usepackage{tikz}
\usepackage{pgfplots}
\usetikzlibrary{bayesnet}
\usepackage{subcaption}
\usetikzlibrary{fit,shapes,arrows,backgrounds,positioning}
\newtheorem{theorem}{Theorem}
\newtheorem{corollary}{Corollary}
\newtheorem{lemma}{Lemma}

\newcommand{\ibraces}[1]{\left\lbrace #1 \right\rbrace_{i \in [L]}}
\newcommand{\braces}[1]{\left\lbrace #1 \right\rbrace}
\newcommand{\parentheses}[1]{\left( #1 \right)}
\newcommand{\jbraces}[1]{\left\lbrace #1 \right\rbrace_{j\in [J]}}

\newcommand{\T}{\mathscr{T}}

\newcommand{\norm}[1]{\left\Vert #1\right\Vert}
\newcommand{\transpose}[1]{\left( #1\right)^T}
\newcommand{\p}[1]{p\left(#1\right)}
\newcommand{\inv}[1]{\left({#1}\right)^{-1}}
\DeclarePairedDelimiter{\ceil}{\lceil}{\rceil}
\DeclareMathOperator*{\argmin}{arg\,min}
\DeclareMathOperator*{\argmax}{arg\,max}
\newtheorem{assumption}{Assumption}
\newcommand{\function}[2]{#1 \left( #2 \right)}
\usepackage{arydshln}

\title{Multimodal Sparse Bayesian Dictionary Learning}

\author{Igor Fedorov, Bhaskar D. Rao

\thanks{I. Fedorov is with the ARM Machine Learning Research Lab, Boston MA. B.D. Rao is with the Department of Electrical and Computer Engineering, University of California, San Diego, 9500 Gilman Dr, San Diego, CA, 92103}}

\begin{document}
\maketitle

\begin{abstract}
This paper addresses the problem
of learning dictionaries for multimodal datasets, i.e. datasets
collected from multiple data sources. We present an algorithm
called multimodal sparse Bayesian dictionary learning (MSBDL).
MSBDL leverages information from all available data modalities
through a joint sparsity constraint. The
underlying  framework offers a considerable amount
of flexibility to practitioners and addresses many of the shortcomings
of existing multimodal dictionary learning approaches. In particular, the procedure
includes the automatic tuning of hyperparameters and is unique in that it allows the dictionaries
for each data modality to have different cardinality, a
significant feature in cases when the dimensionality of data differs
across modalities. MSBDL is scalable and can be used in supervised learning settings.
Theoretical results relating to the convergence
of MSBDL are presented and the numerical results provide evidence
of the superior performance of MSBDL
on synthetic and real datasets compared to existing
methods.

%The purpose of this paper is to address the problem of learning dictionaries for multimodal datasets, i.e. datasets collected from multiple data sources. We present an algorithm called multimodal sparse Bayesian dictionary learning (MSBDL). MSBDL leverages information from all available data modalities through a joint sparsity constraint, displaying superior performance on synthetic and real datasets compared to existing methods. Our approach is unique in that it allows the dictionaries for each data modality to have different cardinality, which is a significant feature in cases when the dimensionality of data differs across modalities. Our framework offers a considerable amount of flexibility to practitioners and addresses many of the shortcomings of existing multimodal dictionary learning approaches, including the automatic tuning of hyperparameters. MSBDL can be used in numerous scenarios, from small datasets to extensive datasets with large dimensionality and in supervised settings. We also report theoretical results relating to the convergence of MSBDL as well as to the benefits of multimodal dictionary learning in general.
\end{abstract}

\section{Introduction}\label{sec:introduction}
Due to improvements in sensor technology, acquiring vast amounts of data has become relatively easy. Given the ability to harvest data, the task becomes how to extract relevant information from the data. The data is often multimodal, which introduces novel challenges in learning from it. Multimodal dictionary learning has become a popular tool for fusing multimodal information \cite{cha2015multimodal,shekhar2014joint,
caicedo2012multimodal,ding2015joint,fedorov2017mutlimodal}. 

Let $L$ and $J$ denote the number of data points for each modality and number of modalities, respectively. Let $Y_j = \begin{bmatrix}
y_j^1 & \cdots & y_j^L
\end{bmatrix} \in \mathbb{R}^{N_j \times L}$ denote the data matrix for modality $j$, where $y_j^i$ denotes the $i$'th data point for modality $j$. We use uppercase symbols to denote matrices and lowercase symbols to denote the corresponding matrix columns. The multimodal dictionary learning problem consists of estimating dictionaries $ D_j \in \mathbb{R}^{N_j \times M_j}$ given data ${Y_j}$ such that 
$
Y_j \approx D_j X_j \; \forall j.
$
We focus on overcomplete dictionaries because they are more flexible in the range of signals they can represent \cite{donoho2006stable}. Since $y_j^i$ admits an infinite number of representations under overcomplete $D_j$, we seek sparse $x_j^i$ \cite{aharon2006ksvd}. 

Without further constraints, the multimodal dictionary learning problem can be viewed as $J$ independent unimodal problems. To fully capture the multimodal nature of the problem, the learning process must be adapted to encode the prior knowledge that each set of points $\bm{y^i} = \jbraces{y_j^i}$ is generated by a common source which is measured $J$ different ways, where $[J] = \braces{1,\cdots,J}$. For any variable $x$, $\bm{x}$ denotes $\jbraces{x_j}$. For instance, in \cite{yang2010image}, low and high resolution image patches are modelled as $y_1^i$ and $y_2^i$, respectively, and the association between $y_1^i$ and $y_2^i$ is enforced by the constraint $x_1^i = x_2^i$. The resulting multimodal dictionary learning problem, referred to here as $\ell_1$DL, is to solve \cite{mairal2009online}
\begin{align}\label{eq:l1}
\argmin_{\tilde{D},X} \norm{{\tilde{Y}} - {\tilde{D}} {X}}_F^2 + & \lambda \norm{X}_1
\\ \nonumber
\tilde{Y}
= \begin{bmatrix}
Y_1^T & \cdots & Y_J^T
\end{bmatrix}^T, & \; \tilde{D} = \begin{bmatrix}
D_1^T & \cdots & D_J^T
\end{bmatrix}^T,
\end{align}
where $\norm{\cdot}_F$ is the Frobenius norm, $\norm{X}_1 = \sum_{i \in [L]} \norm{x^i}_1$, and the $\ell_1$-norm is used as a convex proxy to the $\ell_0$ sparsity measure. In a classification framework, \eqref{eq:l1} can be viewed as learning a multimodal feature extractor, where the optimizer is the multimodal representation of $\bm{ y^i}$ that is fed into a classifier \cite{cha2015multimodal,chaNips,shekhar2014joint}. There are $4$ significant deficiencies associated with using $\ell_1$DL for multimodal dictionary learning. In fact, all existing approaches suffer from one of more of the following\footnote{See Table \ref{table:deficiency}.}:
\begin{enumerate}[label=\textbf{D\arabic*}]
\item \label{D:1}  While using the same sparse code for each modality establishes an explicit relationship between the dictionaries for each modality, the same coefficient values may not be able to represent different modalities well.
\item \label{D:2}  Some data modalities are often less noisy than others and the algorithm should be able to leverage the clean modalities to learn on the noisy ones. Since \eqref{eq:l1} constrains $\lambda$ to be the same for all modalities, it is unclear how the learning algorithm can incorporate prior knowledge about the noise level of each modality.
\item \label{D:3}  The formulation in \eqref{eq:l1} constrains $M_j = M \; \forall j$, for some $M$. Since dimensionality can vary across modalities, it is desirable to allow $M_j$ to vary.
\item \label{D:4} The choice of $\lambda$ is central to the success of approaches like \eqref{eq:l1}. If $\lambda$ is chosen too high, the reconstruction term is ignored completely, leading to poor dictionary quality. If $\lambda$ is chosen too low, the sparsity promoting term is effectively eliminated from the objective function. Extensive work has been done to approach this hyperparameter selection problem from various angles. Two popular approaches include treating the hyperparameter selection problem as an optimization problem of its own \cite{bergstra2012random,bergstra2011algorithms} and grid search, with the latter being the prevailing strategy in the dictionary learning community \cite{mairal2009online,jiang2013label}. In either case, optimization of $\lambda$ involves evaluating \eqref{eq:l1} at various choices of $\lambda$, which can be computationally intensive and lead to suboptimal results in practice. For some of the successful multimodal dictionary learning algorithms \cite{fedorov2017mutlimodal,ding2015joint}, where each modality is associated with at least one hyperparameter, grid search becomes prohibitive since the search space grows exponentially with $J$. For more details, see Section \ref{sec:learning complexity}, specifically Table \ref{table:learning complexity} and Fig. \ref{fig:learning complexity figure}, for a comparison of competing algorithms in terms of computational complexity when grid-search hyperparameter tuning is taken into account.
\end{enumerate}

Next, we review relevant works from the dictionary learning literature, highlighting each method's benefits and drawbacks in light of \ref{D:1}-\ref{D:4}. Table \ref{table:deficiency} summarizes where competing algorithms fall short in terms of \ref{D:1}-\ref{D:4}. 

In past work, $M_j = M \; \forall j$, thus exhibiting \ref{D:3}.  One of the seminal unimodal dictionary learning algorithms is K-SVD, which optimizes \cite{aharon2006ksvd}
\begin{align}
\label{eq:ksvd}
\argmin_{D, \; \ibraces{\norm{ x^i }_0 \leq s }} \norm{Y - D X}_F^2,
\end{align}
where $s$ denotes the desired sparsity level and modality subscripts have been omitted for brevity. The K-SVD algorithm proceeds in a block-coordinate descent fashion, where $D$ is optimized while holding $X$ fixed and vice-versa. The update of $X$ involves a greedy $\ell_0$ pseudo-norm minimization procedure \cite{pati1993orthogonal}. In a multimodal setting, K-SVD can be naively adapted, where $Y$ is replaced by $\tilde{Y}$ and $D$ by $\tilde{D}$ in \eqref{eq:ksvd}, as in \eqref{eq:l1}.

One recent approach, referred to here as Joint $\ell_0$ Dictionary Learning (J$\ell_0$DL), builds upon K-SVD for the multimodal dictionary learning problem and proposes to solve \cite{ding2015joint}
\begin{align}\label{eq:j0dl}
\argmin_{\ibraces{ \jbraces{ \chi_j^i = \chi^i} , \vert \chi^i \vert \leq s}} \sum_{j=1}^{J} \lambda_j \norm{Y_j - D_j X_j}_F^2 
\end{align}
where $\chi_j^i$ denotes the support of $x_j^i$. The J$\ell_0$DL algorithm tackles \ref{D:1}-\ref{D:2} by establishing a correspondence between the supports of the sparse codes for each modality and by allowing modality specific regularization parameters, which allow for encoding prior information about the noise level of $y_j^i$. On the other hand, J$\ell_0$DL does not address \ref{D:3} and presents even more of a challenge than $\ell_1$DL with respect to \ref{D:4} since the size of the grid search needed to find $\bm{\lambda}$ grows exponentially with $J$\footnote{See Section \ref{sec:learning complexity}.}. Another major drawback of J$\ell_0$DL is that, since \eqref{eq:j0dl} has an $\ell_0$ type constraint, solving it requires a greedy algorithm which can produce poor solutions, especially if some modalities are much noisier than others \cite{baron2009distributed}.

The multimodal version of \eqref{eq:l1}, referred to here as Joint $\ell_1$ DL (J$\ell_1$DL), seeks \cite{bahrampour}
\begin{align}
\label{eq:jl1dl}
&\argmin_{\bm{D},\bm{X}} \frac{1}{2} \sum_{i \in [L], j \in [J]} \norm{ y_j^i - D_j x_j^i }_2^2 + \lambda \sum_{i\in [L]} \norm{ \Pi^i}_{12}
\end{align}
where $\Pi^i = \begin{bmatrix}
x_1^i & \cdots & x_J^i
\end{bmatrix}$, $\norm{ \Pi^i }_{12} = \sum_{m \in [M]} \norm{\Pi^i[m,:]}_2,$ and $\Pi^i[m,:]$ denotes the $m$'th row of $\Pi^i$. The $\ell_{12}$-norm in \eqref{eq:jl1dl} promotes row sparsity in $\Pi^i$, which promotes $\bm{x^i}$ that share the same support. Interestingly, regularizers which promote common support amongst variables appear in a wide range of works, including graphical model identification \cite{zorzi2016ar,alpago2018identification}. Like all of the previous approaches, J$\ell_1$DL adopts a block-coordinate descent approach to solving \eqref{eq:jl1dl}, where an alternating direction method of multipliers algorithm is used to compute the sparse code update stage \cite{parikh2014proximal}. While J$\ell_1$DL makes progress toward addressing \ref{D:1}, it does so at the cost of sacrificing the hard constraint that $\bm{x^i}$ share the same support. The authors of \cite{bahrampour} attempt to address \ref{D:2} by relaxing the constraint on the support of $\bm{ x^i } $ even more, but we will not study this approach here because it moves even further from the theme of this work. In addition, J$\ell_1$DL does not address \ref{D:3}-\ref{D:4}. Although \ref{D:4} is not as serious of an issue for J$\ell_1$DL as for J$\ell_0$DL because J$\ell_1$DL requires tuning a single hyperparameter, \ref{D:3} limits the usefulness of J$\ell_1$DL in situations where the observed modalities are drastically different in type and cardinality, i.e. text and images\footnote{See Section \ref{sec:phototweet}.}.

One desirable property of dictionary learning approaches is that they be scalable with respect to the size of the dictionary as well as the dataset. When $L$ becomes large, the algorithm must be able to learn in a stochastic manner, where only a subset of the data samples need to be processed at each iteration. Stochastic learning strategies have been studied in the context of $\ell_1$DL \cite{mairal2009online,mairal2012task} and J$\ell_1$DL \cite{bahrampour}, but not for K-SVD or J$\ell_0$DL. Likewise, the algorithm should be able to accommodate large $N_j$.

\begin{table}
\centering
\resizebox{\columnwidth}{!}{%
\begin{tabular}{|ccc:cccc|}
\hline
 & \multicolumn{2}{c}{MSBDL} & J$\ell_1$DL & J$\ell_0$DL & $\ell_1$DL & KSVD \\
 & Proposed & \cite{fedorov2017mutlimodal} & & & & \\
 \hline
\ref{D:1} & & & & &\checkmark & \checkmark \\
\ref{D:2} & & &\checkmark & &\checkmark & \checkmark\\
\ref{D:3} & & \checkmark& \checkmark&\checkmark &\checkmark & \checkmark\\
\ref{D:4} & & \checkmark & \checkmark &\checkmark &\checkmark & \checkmark\\ 
\hline
\end{tabular}}
\caption{Deficiencies that existing dictionary learning algorithms suffer from.}
\label{table:deficiency}
\end{table}

When the dictionary learning algorithm is to be used as a building block in a classification framework, class information can be incorporated within the learning process. In a supervised setting, the input to the algorithm is $\braces{ \bm{Y} , H }$, where $H = \begin{bmatrix}
h^1 & \cdots & h^L
\end{bmatrix} \in \mathbb{B}^{C \times L}$ is the binary class label matrix for the dataset and $C$ is the number of classes. Each $h^i$ is the label for the $i$'th data point in a one-of-$C$ format. This type of dictionary learning is referred to as task-driven \cite{bahrampour,mairal2012task}, label consistent \cite{jiang2013label}, or discriminative \cite{zhang2010discriminative} and the goal is to learn a dictionary $D_j$ such that $x_j^i$ is indicative of the class label. For instance, discriminative K-SVD (D-KSVD) optimizes \cite{zhang2010discriminative}
\begin{align}\label{eq:DKSVD}
\argmin_{D,W, \ibraces{ \norm{ x^i }_0 \leq s }} \norm{Y-DX}_F^2 + \lambda_{su} \norm{H - W X}_F^2 
\end{align}
where $W$ can be viewed as a linear classifier for $x^i$. 

Task-driven $\ell_1$DL (TD-$\ell_1$DL) optimizes \cite{mairal2012task}
\begin{align}
\argmin_{D,W} E_x \left[l_{su}(h,W,x^*(y,D) \right]+\nu \norm{W}_F^2
\end{align}
where $E_x[\cdot]$ denotes the expectation with respect to $\p{x}$, $l_{su}(\cdot)$ denotes the supervised loss function\footnote{Examples of supervised losses include the squared loss in \eqref{eq:DKSVD}, logistic loss, and hinge loss \cite{mairal2012task}.}, $x^*(y,D)$ denotes the solution of \eqref{eq:l1} with the dictionary fixed to $D$, and the last term provides regularization on $W$ to avoid over-fitting.

Task-driven J$\ell_1$DL (TD-J$\ell_1$DL) optimizes \cite{bahrampour}
\begin{small}
\begin{align*}
\argmin_{\bm{D}, \bm{W}} E_{\bm{x}} \left[ \sum_{j \in [J]} l_{su}(h_j,W_j,x_j^*(y_j,D_j) \right]+\nu \sum_{j \in [J]} \norm{W_j}_F^2
\end{align*}
\end{small}
where $x_j^*(y_j,D_j)$ denotes the $j$'th modality sparse code in the optimizer of \eqref{eq:jl1dl} with fixed dictionaries.

\subsection{Contributions}
\begin{figure}
    \centering
    \includegraphics[scale=0.4]{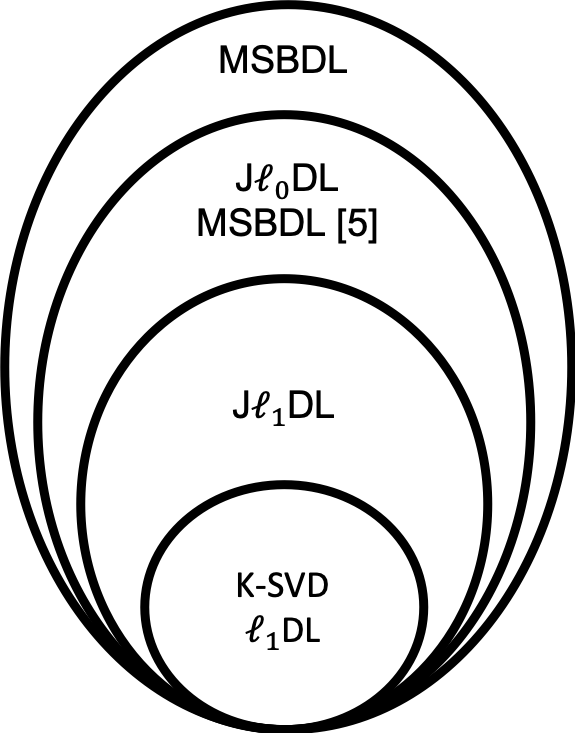}
    \caption{Modeling capacity of proposed algorithm compared to existing works.}
    \label{fig:modeling capacity}
\end{figure}

We present the multimodal sparse Bayesian dictionary learning algorithm (MSBDL). MSBDL is based on a hierarchical sparse Bayesian model, which was introduced in the context of Sparse Bayesian Learning (SBL) \cite{girolami2001variational,tipping2001sparse,wipf2004sparse} as well as dictionary learning \cite{girolami2001variational}, and has since been extended to various structured learning problems \cite{fedorov2016robust,fedorov2016unified,nalci2016rectified}. We presented initial work on this approach in \cite{fedorov2017mutlimodal}\footnote{The algorithm presented in \cite{fedorov2017mutlimodal} was also called MSBDL and we distinguish it from the algorithm presented here by appending the appropriate citation to the algorithm name whenever referencing our previous work.}, where we made some progress towards tackling \ref{D:1}-\ref{D:2}. Here, we go beyond our preliminary work in a number of significant ways, which are summarized in Table \ref{table:comparison with icassp}. We address \ref{D:1}-\ref{D:2} to a fuller extent by offering scalable and task-driven variants of MSBDL. More importantly, we tackle \ref{D:3}-\ref{D:4}, where our solution to \ref{D:4} is crucial to the ability of MSBDL to address \ref{D:1} and resolves a major hyperparameter tuning issue in \cite{fedorov2017mutlimodal}. By  resolving \ref{D:3}, the present work represents a large class of algorithms that contains \cite{fedorov2017mutlimodal} as a special case. This point is depicted in Fig. \ref{fig:modeling capacity}, which provides a visualization of the types of models and variable interactions that can be learned by each of the competing algorithms.  

In summary:
%where we showed that MSBDL is able to capture the relationship between diverse datasets within the dictionary learning process, while addressing \ref{D:1}-\ref{D:2} and benefiting from the significant performance gains afforded by the Bayesian model \cite{giri2015type,ding2015joint,ji2009multitask}. Here, we extend MSBDL to tackle \ref{D:1}-\ref{D:4} as well as scalable and task-driven learning:
\begin{enumerate}
\item We extend MSBDL to address \ref{D:3}. To the best of our knowledge, MSBDL is the first dictionary learning algorithm capable of learning differently sized dictionaries.
\item We extend MSBDL to the task-driven learning scenario. 
\item We present scalable versions of MSBDL. 
\item We optimize algorithm hyperparameters during learning, obviating the need for grid search, and conduct a theoretical analysis of the approach.
\item We show that multimodal dictionary learning offers provable advantages over unimodal dictionary learning.
\end{enumerate}

\begin{table}
\centering
\begin{tabular}{|ccc|}
\hline
 & Proposed & \cite{fedorov2017mutlimodal} \\ \hline
Scalable & \checkmark & \\
Task-driven extension & \checkmark & \\
Solves \ref{D:3} & \checkmark & \\
Solves \ref{D:4} & \checkmark & \\ 
\hline
\end{tabular}
\caption{Comparison of proposed MSBDL variant with \cite{fedorov2017mutlimodal}.}
\label{table:comparison with icassp}
\end{table}

\subsection{On the Motivation of this Work}
\label{sec:motivation}
\begin{figure}
    \centering
    \includegraphics[scale=0.2]{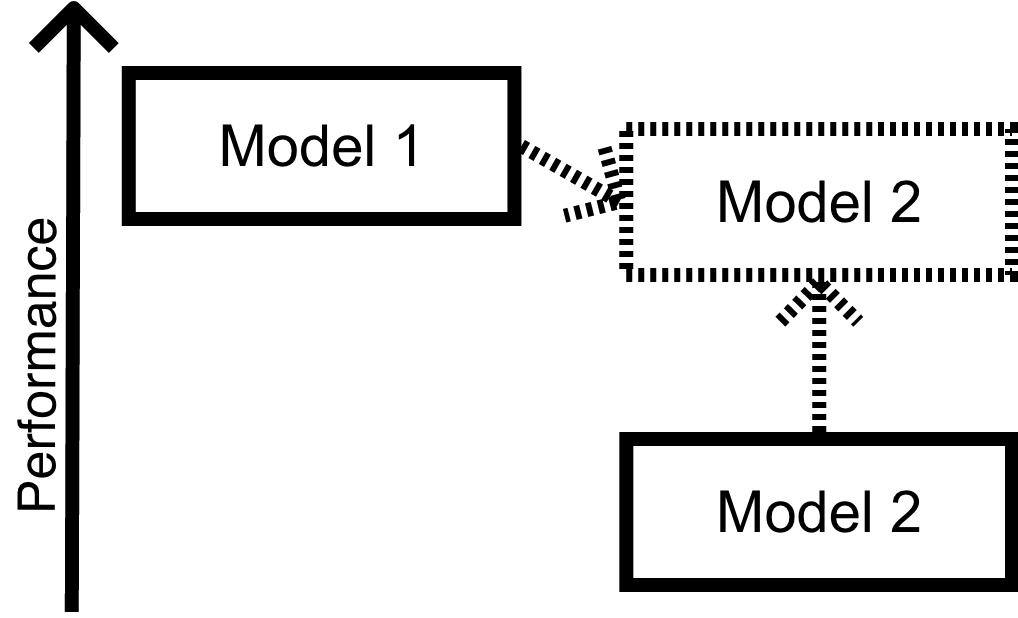}
    \caption{Visualization of high level problem motivation. }
    \label{fig:motivation}
\end{figure}

Consider the scenario depicted in Fig. \ref{fig:motivation}. The solid rectangles represent learning models 1 and 2, corresponding to modalities 1 and 2, respectively, independently of each other. The intuition behind this work is that model 2 can be improved by performing learning in a joint fashion, where information flows between models 1 and 2, as depicted by the dotted shapes in Fig. \ref{fig:motivation}. The experimental and theoretical results reported in this paper all serve as evidence for the ability of MSBDL to leverage all available data sources to learn models for each modality that are superior to ones that could be found if learning was done independently. As such, the aims of this work are, in some sense, tangential to works like \cite{bahrampour}, where the goal is more to learn an information fusion scheme and less to learn superior models for each modality. 

\section{Proposed Approach}
\label{sec:MSBDL}
\begin{figure}
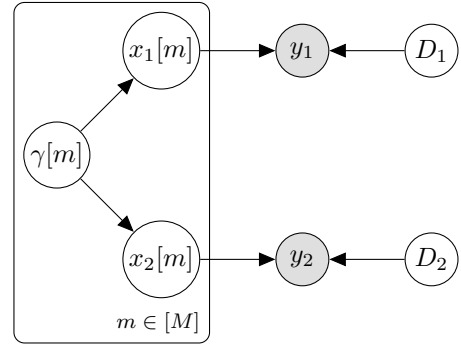

\centering
\tikz{
\node[latent](g1){$\gamma[m]$};
\node[latent,above right=of g1] (x1) {$x_{1}[m]$};
\node[obs,right=of x1](y1){$y_{1}$};
\node[latent,right=of y1] (D1) {$D_{1}$};

%\node[latent,below=of g1](g2){$\gamma_2$};
\node[latent,below right=of g1] (x2) {$x_{2}[m]$};
\node[obs,right= of x2](y2){$y_{2}$};
\node[latent,right=of y2] (D2) {$D_{2}$};

\plate{}{(g1)(x1)(x2)}{$m \in [M]$};

%\edge[-]{g1}{g2};
\edge[->]{g1}{x1};
\edge[->]{g1}{x2};
\edge[->]{x1}{y1};
\edge[<-]{y1}{D1};
\edge[->]{x2}{y2};
\edge[<-]{y2}{D2};
}
\caption{Graphical model for two modality MSBDL.}
\label{fig:graphical model msbdl}
\end{figure}

The graphical model for MSBDL is shown in Fig. \ref{fig:graphical model msbdl}. The signal model is given by 
\begin{align}\label{eq:signal model}
{y_j} = {D_j x_j}+{v_j}, \; \; {v_j} \sim \mathsf{N}\parentheses{0,\sigma_j^2 {\mathsf{I}}}
\end{align}
where $\mathsf{N}(\cdot)$ denotes a Gaussian distribution and the noise variance is allowed to vary among modalities. In order to promote sparse $x_j^i$, we assume a Gaussian Scale Mixture (GSM) prior on each element of ${x_j}$ \cite{tipping2001sparse,wipf2004sparse,wipf2007empirical,wipf2011latent}. The GSM prior is a popular class of distributions whose members include many sparsity promoting priors, such as the Laplacian and Student's-t \cite{tipping2001sparse,wipf2004sparse,wipf2007empirical,wipf2011latent,
palmer2006variational,giri2015type}. The remaining task is to specify the conditional density of ${x_j}$ given ${\gamma}$. One option is to use what we refer to as the one-to-one prior \cite{fedorov2017mutlimodal}:
\begin{align}\label{eq:gsm}
\p{x_j | \gamma} = \prod_{m \in [M]} \p{x_j[m]|\gamma[m]} = \prod_{m \in [M]} \function{\mathsf{N}}{0,\gamma[m]}
\end{align} 
where $x_j[m]$ denotes the $m$'th element of $x_j$ and the choice of $p(\gamma[m])$ determines the marginal density of ${x_j[m]}$. We assume a non-informative prior on ${\gamma_j[m]}$ \cite{tipping2001sparse}. As will be shown in Section \ref{sec:joint sparsity}, the conditional distribution in \eqref{eq:gsm} represents a Bayesian realization of the constraint that $\bm{x}$ share the same support. The prior in \eqref{eq:gsm} still constrains $M_j = M \; \forall j$, but this restriction will be lifted in Section \ref{sec:modeling complex relationships}. When $J=1$, this model is identical to the one used in \cite{girolami2001variational}.
\vspace{-1em}
\subsection{Inference Procedure}
\label{sec:msbdl inference}
We adopt an empirical Bayesian inference strategy to estimate $\theta = \braces{ \bm{D} , \ibraces{ \gamma^i }}$ by optimizing \cite{mackay1992bayesian}
\begin{align}\label{eq:ML}
\argmax_\theta & \log \p{\bm{Y} | \theta} = \argmax_\theta \sum_{j \in [J]} \log \p{{Y_j} | \theta}\\
\label{eq:log likelihood}
\p{\bm{Y} | \theta} &= \prod_{i \in [L], j \in [J]} \p{y_j^i | \theta}, \p{y_j^i | \theta} = \mathsf{N}(0,\Sigma_y^i) \\
\label{eq:Sigma y}
\Sigma_{y,j}^i &= \sigma_j^2 \mathsf{I}+D_j \Gamma^i D_j^T, \; \; 
\Gamma^i = \text{diag}(\gamma^i).
\end{align}
We use Expectation-Maximization (EM) to maximize \eqref{eq:ML}, where $\braces{ \bm{X},\bm{Y},\theta }$ and $\bm{X}$ are treated as the complete and nuisance data, respectively \cite{dempster1977maximum}. At iteration $t$, the E-step computes $\function{Q}{\theta,\theta^t} = \left \langle \log \p{\bm{Y},\bm{X},\bm{D},\ibraces{ \gamma^i}} \right\rangle$, where $\langle \cdot \rangle$ denotes the expectation with respect to $\p{ \bm{X}  | \bm{Y},\theta^t}$, and $\theta^t$ denotes the estimate of $\theta$ at iteration $t$.  Due to the conditional independence properties of the model, the posterior factors over $i$ and 
\begin{align}\label{eq:posterior}
\function{p}{x_j^i | y_j^i,\theta} &= \function{\mathsf{N}}{\mu_j^i,\Sigma_{x,j}^i} \\
{\Sigma}_{x,j}^i &= \inv{\sigma_j^{-2} {D}_j^T {D}_j + \inv{\Gamma^i}} \label{eq:Sigma update} \\
{\mu}_j^i &= \sigma_j^{-2} {\Sigma}_{x,j}^i {D}_j^T y_j^i \label{eq:mu update}.
\end{align}
In the M-step, $\function{Q}{\theta,\theta^t}$ is maximized with respect to $\theta$. In general, the M-step depends on the choice of $\p{\bm{x} | \gamma}$. For the choice in \eqref{eq:gsm}, the M-step becomes
\begin{align}
\label{eq:gamma update block}
\left(\gamma^i[m]\right)^{t+1} &= J^{-1} {\sum_{j=1}^J {\Sigma}_{x,j}^{i} [m,m] + \left({\mu}_j^{i} [m]\right)^2}\\
D_j^{t+1} &= Y_j  {U}_j^T \inv{{U}_j {U}_j^T+\sum_{i \in [L]} {\Sigma}_{x,j}^{i}}\label{eq:D update} \\
U_j &= \begin{bmatrix}
\mu_j^1 & \cdots & \mu_j^L
\end{bmatrix}.
\end{align}

\subsection{How does MSBDL solve deficiency \ref{D:1}?}
\label{sec:joint sparsity}
One consequence of the GSM prior is that many of the elements of $\gamma^i$ converge to $0$ during inference \cite{wipf2004sparse}. When $\gamma^i[m] = 0$, $\p{x_j^i[m] | y_j^i,\gamma^i}$ reduces to $\delta\left(x_j[m] \right)$ for all $j$, where $\delta(\cdot)$ denotes the Dirac-delta function \cite{wipf2004sparse}. Since the only role of $\bm{x}$ in the inference procedure is in the E-step, where we take the expectation of the complete data log-likelihood with respect to $\p{x_j^i | y_j^i,\gamma^i}$, the effect of having $\p{x_j^i[m] | y_j^i,\gamma^i} = \function{\delta}{x_j^i[m]}$ is that the E-step reduces to evaluating the complete data log-likelihood at $x_j^i[m] = 0, \forall j$. Therefore, upon convergence, the proposed approach exhibits the property that $\bm{x^i}$ share the same support.

\subsection{Connection to J$\ell_1$DL}
Suppose that the conditional distribution in \eqref{eq:gsm} is used with $
{\gamma[m]} \sim \mathsf{Ga}\left({J}/{2},1\right) \; \forall m
$,
where $\mathsf{Ga}(\cdot)$ refers to the Gamma distribution. It can be shown that \cite{boisbunonclass}
\begin{align*}
p\left(\bm{x^i} \right) = c \prod_{m \in [M]} \function{K_0}{ \norm{ \begin{bmatrix}
x_1^i[m] & \cdots & x_J^i [m]
\end{bmatrix} }_{2}},
\end{align*}
where $c$ is a normalization constant and $K_0(\cdot)$ denotes the modified Bessel function of the second kind and order $0$. For large $x$, $K_0(x) \approx {\pi} \exp(-x) / {\sqrt{2\pi x}}$ \cite{eltoft2006multivariate}. In the following, we replace $K_0(x)$ by its approximation for purposes of exposition. Under the constraint $\sigma_j = 0.5 \lambda \; \forall j$, the MAP estimate of $\braces{ \bm{D} , \bm{X} }$ is given by
\begin{align}\label{eq:type 1 msbdl}
\begin{split}
\argmin_{\bm{D},\bm{X}} & \sum_{i \in [L],j \in [J]} \norm{ y_j^i - D_j x_j^i }_2^2 + \lambda \sum_{i \in [L]} \norm{ \Pi^i}_{12} + \\& 0.5\lambda \sum_{i \in [L],m \in [M]} \log  \norm{ \Pi^i[m,:] }_2.
\end{split}
\end{align}
This analysis exposes a number of similarities between MSBDL and J$\ell_1$DL. If we ignore the last term in \eqref{eq:type 1 msbdl}, J$\ell_1$DL becomes the MAP estimator of $\braces{\bm{D},\bm{X}}$ under the one-to-one prior in \eqref{eq:gsm}. If we keep the last term in \eqref{eq:type 1 msbdl}, the effect is to add a Generalized Double Pareto prior on the $\ell_2$ norm of the rows of $\Pi^i$ \cite{giri2015type,c5}. 

At the same time, there are significant differences between MSBDL and J$\ell_1$DL. The J$\ell_1$DL objective function assumes ${\sigma_j}$ is constant across modalities, which can lead to a strong mismatch between data and model when the dataset consists of sources with disparate noise levels. In contrast to J$\ell_1$DL, MSBDL enjoys the benefits of evidence maximization \cite{mackay1996hyperparameters,mackay1992bayesian,giri2015type}, which naturally embodies "Occam's razor" by penalizing unnecessarily complex models and searches for areas of large posterior mass, which are more informative than the mode when the posterior is not well-behaved. 

\section{Complete Algorithm}
\label{sec:complete msbdl algorithm}
So far, it has been assumed that $\bm{\sigma}$ is known. Although it is possible to include $\bm{\sigma}$ in $\theta$ and estimate it within the evidence maximization framework in \eqref{eq:ML}, we experimental observe that $\bm{\sigma}$ decays very quickly and the algorithm tends to get stuck in poor stationary points. An alternative approach is described next. Consider the scenario where $D_j$ is known and we seek the MAP estimate of $x_j$ given $y_j,D_j$:
\begin{align}\label{eq:map}
\argmin_{x_j} \norm{y_j - D_j x_j}_2^2 - 2\sigma_j^2\log \p{x_j}.
\end{align}
The estimator in \eqref{eq:map} shows that $\sigma_j$ can be thought of as a regularization parameter which controls the trade-off between sparsity and reconstruction error. As such, we propose to anneal $\sigma_j$. The motivation for annealing $\sigma_j$ is that the quality of $D_j$ increases with $t$, so giving too much weight to the reconstruction error term early on can force EM to converge to a poor stationary point. 

Let $\sigma_j^0 > \sigma^\infty \geq 0,\alpha_\sigma < 1$, $\tilde{\sigma}_j^{t+1} = \max(\sigma^\infty,\alpha_\sigma \sigma_j^t)$. The proposed annealing strategy can then be stated as
\begin{small}
\begin{align}\label{eq:sigma update}
\sigma_j^{t+1} = 
\begin{cases}
\tilde{\sigma}_j^{t+1} & \text{ if } \log \p{Y_j | \theta^{t+1},\tilde{\sigma}_j^{t+1}} > \log \p{Y_j | \theta^{t+1},\sigma_j^t} \\
\sigma_j^t & \text{ else. }
\end{cases}
\end{align}
\end{small}
Although it may seem that we have replaced the task of selecting $\bm{\sigma}$ with that of selecting $\braces{\bm{\sigma^0}, \alpha_\sigma, \sigma^\infty}$, we claim that the latter is easy to select and provide both theoretical (Section \ref{sec:analysis}) and experimental (Section \ref{sec:results}) validation for this claim. The main benefit of the proposed approach is that it essentially traverses a continuous space of candidate $\bm{\sigma}$ without explicitly performing a grid search, which would be intractable as $J$ grows. The parameter $\sigma^\infty$ can be set arbitrarily small and $\bm{\sigma^0}$ can be set arbitrarily large. The only recommendation we make is to set $\sigma_j^0 > \sigma_{j'}^0$ if modality $j$ is a-priori believed to be more noisy than modality $j'$.

Section \ref{sec:analysis} studies the motivation for and properties of the annealing strategy in greater detail. In practice, the computation of $\p{Y_j | \theta^{t+1},\tilde{\sigma}_j^{t+1}}$ is costly because it requires the computation of the sufficient statistics in \eqref{eq:Sigma y} for all $L$ data points and $J$ modalities. Instead, we replace the condition in \eqref{eq:sigma update} by checking whether decreasing $\sigma_j$ should increase $\p{Y_j | \theta^{t+1},\sigma^t}$. This check is performed by checking the sign of the first derivative of $\p{Y_j | \theta^{t+1},\sigma^t}$, replacing \eqref{eq:sigma update} with
\begin{align}\label{eq:sigma update smart}
\sigma_j^{t+1} = 
\begin{cases}
\tilde{\sigma}_j^{t+1} & \text{ if } {\partial \log \p{Y_j | \theta^{t+1},\sigma_j^t}}/{\partial \sigma_j^t} < 0 \\
\sigma_j^t & \text{ else. }
\end{cases}
\end{align}
It can be shown that \eqref{eq:sigma update smart} can be computed essentially for free by leveraging the sufficient statistics computed in the $\theta$ update step\footnote{See Supplemental Material.}. The complete MSBDL algorithm is summarized in Fig. \ref{alg:MSBDL complete}. In practice, each ${D_j}$ is normalized to unit $\ell_2$ column norm at each iteration to prevent scaling instabilities. 

\begin{figure}
\begin{algorithmic}[1]
\Require $\bm{Y},\bm{\sigma^0},\sigma^\infty,\alpha_\sigma$
\While{$\bm{\sigma}$ not converged}
\While{$\bm{D}$ not converged}
\For{$i \in [L]$}
\State Update $\bm{\Sigma_x^i}, \bm{\mu^i}, \gamma^i$ using \eqref{eq:Sigma update}, \eqref{eq:mu update}, and \eqref{eq:gamma update block}
\EndFor
\State $\jbraces{ \text{ Update } D_j \text{ using \eqref{eq:D update} if } \sigma_j \text{ not converged}}$
\EndWhile 
\State $\jbraces{ \text{Update $\sigma_j$ using \eqref{eq:sigma update smart} if $\sigma_j$ not converged} }$
\EndWhile
\end{algorithmic}
\caption{MSBDL algorithm for the one-to-one prior in \eqref{eq:gsm}.}
\label{alg:MSBDL complete}
\end{figure}

\subsection{Dictionary Cleaning}
We adopt the methodology in \cite{aharon2006ksvd} and ``clean" each ${D_j}$ every $T$ iterations. Cleaning $D_j$ means removing atoms which are aligned with one or more other atoms and replacing the removed atom with the element of $Y_j$ which has the poorest reconstruction under $D_j$. A given atom is also replaced if it does not contribute to explaining $Y_j$, as measured by the energy of the corresponding row of $U_j$.

\section{Scalable Learning}
When $L$ is large, it is impractical to update the sufficient statistics for all data points at each EM iteration. To draw a parallel with stochastic gradient descent (SGD), when the objective function is a sum of functions of individual data points, one can traverse the gradient with respect to a randomly chosen data point at each iteration instead of computing the gradient with respect to every sample in the dataset. In the dictionary learning community, SGD is often the optimization algorithm of choice because the objective function is separable over each data point \cite{mairal2009online,mairal2012task,bahrampour}. In addition, as $N_j$ grows, the computation of the sufficient statistics in \eqref{eq:Sigma update}-\eqref{eq:mu update} can become intractable. In the following, we propose to address these issues using a variety of modifications of the MSBDL algorithm from Section \ref{sec:MSBDL}. The proposed methods also apply to priors other than the one in \eqref{eq:gsm} \footnote{See Section \ref{sec:modeling complex relationships}}.

\subsection{Scalability with respect to the size of the dataset}
In the following, we present two alternatives to the EM MSBDL algorithm to achieve scalability with respect to $L$. The first proposed approach, referred to here as Batch EM, computes sufficient statistics only for a randomly chosen subset $\phi = \lbrace i^1,\cdots,i^{L_0} \rbrace$ at each EM iteration, where $L_0$ denotes the batch size. The M-step consists of updating $\lbrace \gamma^i \rbrace_{i \in \phi}$ using \eqref{eq:gamma update block} and updating $D_j$ using \eqref{eq:D update}, with the exception that only the sufficient statistics from $i \in \phi$ are employed.

Another stochastic inference alternative is called Incremental EM, which is reviewed in the Supplementary Material \cite{neal1998view}. In the context of MSBDL, Incremental EM is tantamount to an inference procedure which, at each iteration, randomly selects a subset $\phi$ of points and updates the sufficient statistics in \eqref{eq:Sigma update}-\eqref{eq:mu update} for $i \in \phi$. During the M-step, the hyperparameters $\lbrace \gamma^i \rbrace_{i \in \phi}$ are updated. The dictionaries $D_j$ are updated using \eqref{eq:D update}, where the update rule depends on sufficient statistics computed for all $L$ data points, only a subset of which have been updated during the given iteration. 

\begin{table}
\centering
\begin{tabular}{ccccc}
	& EM Type & $\mu/\Sigma_x$ & CC & MC\\ \hline
MSBDL   & Exact       & Exact & $L N^3$     & $L M^2$ \\
MSBDL-1 & Incremental & Exact & $L_0 N^3$ & $L M^2$ \\
MSBDL-2 & Incremental & Approximate & $L_0 N^2 \ceil{N} $ & $LM$\\
MSBDL-3 & Batch & Exact & $L_0 N^3$ & $L_0 M^2$\\
MSBDL-4 & Batch & Approximate & $L_0 N^2 \ceil{N}$ & $L_0 M$
\end{tabular}
\caption{CC, MC, $L_0$, and $\ceil{N}$ denote worst case computational complexity, worst case memory complexity, batch size, and a quantity which is upper-bounded by $N$, respectively.}
\label{table:taxonomy}
\end{table}

% \begin{table}
% \centering
% \resizebox{\columnwidth}{!}{%
% \begin{tabular}{|ccccc|}
% \hline
% MSBDL & J$\ell_1$DL & J$\ell_0$DL & $\ell_1$DL & KSVD \\ \hline
% $T L N^3 J $ & $F TL M^3 J^3$ & $F^J T L J M N s$ & $F T L (M^3 + M^2 N J)$ \cite{efron2004least} & $T L (M^3 + M^2 N J)$\cite{rakotomamonjy2011surveying} \\
% \hline
% \end{tabular}}
% \caption{Learning complexity, where $T$ denotes the number of iterations over the entire training set and $F$ denotes the grid spacing for a grid search over the algorithm hyperparameters. We assume that the hyperparameters consist of $\lambda$ in \eqref{eq:l1} and \eqref{eq:jl1dl} and $\jbraces{\lambda_j}$ in \eqref{eq:j0dl}.}
% \label{table:learning complexity}
% \end{table}

\subsection{Scalability with respect to the size of the dictionary}
%\subsubsection{Diagonal covariance approximation}
In order to avoid the inversion of the $N_j \times N_j$ matrix\footnote{Due to the matrix inversion lemma, \eqref{eq:Sigma update} can be computed using ${\Gamma^i}-\Gamma^i D_j^T\inv{\sigma_j^2 \mathsf{I} + D_j \Gamma^i D_j^T }D_j \Gamma^i$.} required to compute \eqref{eq:Sigma update}, we use the conjugate gradient algorithm to compute $\mu_j^i$ and approximate $\Sigma_{x,j}^i$ by
\begin{align}\label{eq:Sigma approx}
\Sigma_{x,j}^i \approx \inv{diag\left(\sigma^{-2} D_j^T D_j + \inv{\Gamma^i}\right)}
\end{align}
where, in this case, $diag(\cdot)$ denotes setting the off-diagonal elements of the input to $0$ \cite{babacan2011variational,marnissi2017variational}. 

Table \ref{table:taxonomy} shows a taxonomy of MSBDL algorithms considered using exact or incremental EM and exact or approximate computation of \eqref{eq:Sigma update}-\eqref{eq:mu update}, along with the corresponding worst case computational and memory complexity per EM iteration. The Supplementary Material provides a visualization of the difference between the proposed algorithms.

\subsection{Learning complexity}
\label{sec:learning complexity}
\begin{figure}
    \centering
    \includegraphics[scale=0.25]{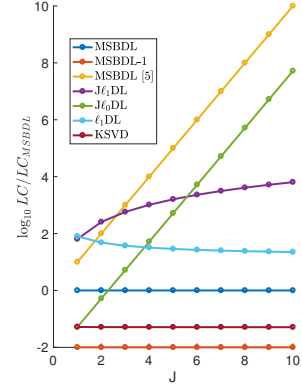}
    \caption{Learning complexity}
    \label{fig:learning complexity figure}
\end{figure}

Finally, we consider the learning complexity (LC) of MSBDL, specifically with respect to that of competing approaches, where we define LC to be the (worst-case) computational complexity of running the algorithm and tuning its hyperparameters. In the case of MSBDL, the hyperparameters are learned jointly with the parameters $\theta$, obviating the need for hyperparameter optimization. In the case of J$\ell_1$DL, J$\ell_0$DL, $\ell_1$DL, and K-SVD, we assume that a grid search is performed to estimate the hyperparameters, with a grid spacing of $F$ for each hyperparameter. Analytical results for LC are shown in Table \ref{table:learning complexity}\footnote[3]{The dependence of LC for J$\ell_1$DL on $(MJ)^3$ comes from the need to solve the least squares problem in [(21) , \cite{bahrampour}], with a matrix of size $MJ \times MJ$, in the worst case.} and Fig. \ref{fig:learning complexity figure} plots the ratio of LC for competing approaches to that of MSBDL as a function of $J$ for $N = 100$, $M=400$, $F=10$, $s=10$, $T=L=1$. For comparison, we also plot the LC of MSBDL-1 with $\ceil{N} = 1$. The results show that J$\ell_0$DL and the MSBDL variant in \cite{fedorov2017mutlimodal} scale least well with $J$, which can be explained by the fact that both require grid-search to tune $J$ hyperparameters. J$\ell_1$DL exhibits worse LC behavior than MSBDL, with the gap between the two methods growing as $J$ increases. Interestingly, MSBDL-1 exhibits the best LC amongst all of the algorithms\footnote{The assumption that $\ceil{N} = 1$ may not be true in practice, so the plot of LC for MSBDL-1 represents the best-case scenario.}. The analysis in this section shows that \ref{D:4} is a serious concern for several existing approaches, especially \cite{fedorov2017mutlimodal}, which motivates the current approach.

\begin{table}
\centering
\begin{tabular}{|cc|}
\hline
MSBDL (proposed) & $T L N^3 J $\\
MSBDL \cite{fedorov2017mutlimodal} & $F^J T L N^3 J $ \\
J$\ell_1$DL & $F T L M^3 J^3 $ \cite{bahrampour} \\
J$\ell_0$DL & $F^J T L J (N (M+s+s^2) + s^3)$ \cite{sturm2012comparison,tropp2005simultaneous}\\
$\ell_1$DL & $F T L (M^3 + M^2 N J)$ \cite{efron2004least} \\
K-SVD & $F T L ( J N (M + s +s^2) + s^3)$\cite{sturm2012comparison} \\ \hline
\end{tabular}
\caption{Learning complexity, where $T$ denotes the number of iterations over the entire training set and $F$ denotes the grid spacing for a grid search over the algorithm hyperparameters. We assume that the hyperparameters consist of $\lambda$ in \eqref{eq:l1} and \eqref{eq:jl1dl}, $\jbraces{\lambda_j}$ in \eqref{eq:j0dl}, and $s$ in \eqref{eq:ksvd}.}
\label{table:learning complexity}
\end{table}

\section{Modeling More Complex Relationships}
\label{sec:modeling complex relationships}
The drawback of the one-to-one prior in \eqref{eq:gsm} is that
%, like all of the methods surveyed in Section \ref{sec:introduction}, 
it constrains $M_j = M \; \forall j$. In the following, we propose two models which allow for $M_j$ to be modality-dependent. For ease of exposition, we set $J=2$, but the models we describe can be readily expanded to $J > 2$. We propose to organize $\bm{x}$ into a tree with $K$ disjoint branches. We adopt the convention that the elements of $x_1$ and $x_2$ form the roots and leaves of the tree, respectively\footnote{We adopt this convention without loss of generality since the modalities can be re-labeled arbitrarily.}. The root of the $k$'th branch is $x_1[k]$ and the leaves are indexed by $\mathscr{T}^k \subseteq [M_2]$. The defining property of the models we propose is the relationship between the sparsity pattern of the root and leaf levels.
%As will be shown, the proposed models lend themselves to closed-form inference.

\subsection{Atom-to-subspace sparsity}
\label{sec:subspace2subspace}

\begin{figure}
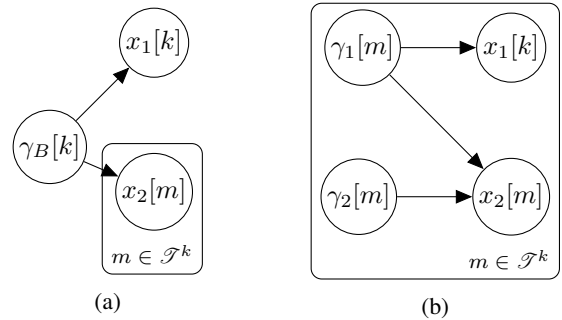

\begin{subfigure}{0.45\columnwidth}
\centering
\tikz{
\node[latent] (x11) {$x_{1}[k]$};
\node[latent, below=of x11] (x21) {$x_2[m]$};
\plate{groupx21}{(x21)}{$m \in \mathscr{T}^k$};
\node[latent,below left=of x11](g){$\gamma_B[k]$};
%\edge[-]{x11}{x21};
\edge[->]{g}{x11,x21}; 
}
\caption{}
\label{fig:atom2subspace}
\end{subfigure}
~
\begin{subfigure}{0.45\columnwidth}
\centering
\tikz{
\node[latent] (x11) {$x_{1}[k]$};
\node[latent, below=of x11] (x21) {$x_2[m]$};

\node[latent,left=of x11](g1){$\gamma_1[m]$};
\node[latent,left=of x21](g2){$\gamma_2[m]$};

\plate{groupx21}{(x21)(g2)(x11)}{$m \in \mathscr{T}^k$};

\edge[->]{g2}{x21};
%\edge[-]{x11}{x21};
\edge[->]{g1}{x11,x21};
}
\caption{}
\label{fig:hierarchical}
\end{subfigure}
\caption{Prototype branches for the atom-to-subspace (\ref{fig:atom2subspace}) and hierarchical sparsity (\ref{fig:hierarchical}) models.}
\end{figure}

%\begin{figure}
%\centering
%\tikz{
%\node[latent] (x11) {$x_{1}[k]$};
%\node[ below=of x11] (dots){$\cdots$};
%\node[latent, left=of dots] (x21) {$x_2\left[m^1\right]$};
%\node[latent, right=of dots] (x22) {$x_2\left[m^{\vert \mathscr{T}^k \vert}\right]$};
%\edge[-]{x11}{x21,x22}; 
%}
%\caption{Prototype branch for proposed class of models.}
%\label{fig:tree model}
%\end{figure}

The one-to-one prior in \eqref{eq:gsm} can be viewed as linking the one-dimensional subspaces spanned by $d_1^m$ and $d_2^m$ for $m \in [M]$. Whenever $d_1^m$ is used to represent $y_1$, $d_{2}^m$ is used to represent $y_{2}$, and vice-versa. The extension to the multi-dimensional subspace case stipulates that if $d_1^k$ is used to represent $y_1$, then $\left \lbrace d_2^m \right \rbrace_{m \in \mathscr{T}^k}$ is used to represent $y_2$, and vice-versa. This model
does not constrain $\vert \mathscr{T}^k \vert$ to be the same for all $k$, such that $M_2$ can be chosen independently of $M_{1}$.

Let ${\gamma_B} \in \mathbb{R}_+^{K}$, in contrast to Section \ref{sec:MSBDL} where $\gamma \in \mathbb{R}_+^M$. We encode the atom-to-subspace sparsity prior by assigning a single hyperparameter $\gamma_B[k]$ to each branch $k$. The distribution $\p{\bm{x} | \gamma_B}$ is then given by\footnote{See Supplementary Material for a derivation of the resulting prior on $\bm{x}$.}
\begin{align}\label{eq:prior block}
\prod_{k \in [K]} \function{\mathsf{N}}{x_1[k];0,\gamma_B[k] }  \prod_{m \in \mathscr{T}^k} \function{\mathsf{N}}{x_2[m];0,\gamma_B[k]}.
\end{align}
Fig. \ref{fig:atom2subspace} shows a prototype branch under the atom-to-subspace prior. Inference for the prior in \eqref{eq:prior block} proceeds in much the same way as in Section \ref{sec:msbdl inference}. The form of the marginal likelihood in \eqref{eq:log likelihood} and posterior in \eqref{eq:posterior} remain the same, with the exception that $\Sigma_{y,j}^i$ and $\Sigma_{x,j}^i$ are re-defined to be
\begin{align}
\Sigma_{y,j}^i &= \sigma_j^2 \mathsf{I}+ D_j \Gamma_j^i D_j^T \label{eq:sigmay block} \\
\Sigma_{x,j}^i &= \inv{\sigma_j^{-2} D_j^T D_j + \inv{\Gamma_j^i}}, \Gamma_1^i = diag(\gamma_B^i) \nonumber,
\end{align}
where $\Gamma_2^i$ is a diagonal matrix whose $[m,m]$'th entry is $\gamma^i_B[k]$ for $m \in \mathscr{T}^k$. The update of $\gamma_{B}^i$ is given by
\begin{align*}
\frac{\Sigma_{x,1}^i[m,m] + \left(\mu_{1}^i[m]\right)^2 +\sum_{m \in \mathscr{T}^k} \Sigma_{x,2}^i[m,m] + \left(\mu_2^i[m]\right)^2}{1+ \vert \mathscr{T}^k \vert},
\end{align*}
while the update of $D_j$ remains identical to \eqref{eq:D update}. There are many ways to extend the atom-to-subspace prior for $J > 2$, depending on the specific application. One possibility is to simply append more branches to the root at $x_1[k]$ corresponding to coefficients from modalities $j > 2$\footnote{See Section \ref{sec:mirflickr}.}.

One problem is that, for $\vert \mathscr{T}^k \vert > 1$, the atoms of $D_2$ indexed by $\mathscr{T}^k$ are not identifiable. The reason for the identifiability issue is that $D_2$ appears in the objective function in \eqref{eq:log likelihood} only through the $D_2 \Gamma_2 D_2^T$ term in \eqref{eq:sigmay block}, which can be written as
\begin{align}
D_2 \Gamma_2 D_2^T = \sum_{k \in [K]} \gamma_B[k] \sum_{m \in \mathscr{T}^k} d_2^m \transpose{d_2^m}.
\end{align}
Therefore, any $D_{2}^{'}$ which satisfies $\sum_{m \in \mathscr{T}^k} {d'}_2^m \transpose{{d'}_2^m} = \sum_{m \in \mathscr{T}^k} d_2^m \transpose{d_2^m}$ for all $k$ achieves the same objective function value as $D_2$. Since the objective function is agnostic to the individual atoms of $D_2$, the performance of this model is severely upper-bounded in terms of the ability to recover $D_2$. In the following, we propose an alternative model which circumvents the identifiability problem.

\subsection{Hierarchical Sparsity}
\label{sec:hierarchical sparsity main}
In this section, we propose a model which allows the root of each branch to control the sparsity of the leaves, but not vice-versa.
%An alternative to the atom-to-subspace model in Section \ref{sec:subspace2subspace} 
%In this section, we will continue to use the grouping defined in Section \ref{sec:subspace2subspace} above, whereby the atoms of $D_j$ are partitioned into disjoint sets indexed by $\left \lbrace \psi_j^k \right\rbrace_{k=1}^K$. For ease of exposition let $J=2$. 
%We begin by considering the case where $M_1 < M_2$, $\vert \psi_1^k \vert = 1 \forall k$, and $\vert \psi_2^k \vert = 2 \forall k$. Suppose that the coefficients of $x_1$ and $x_2$ are organized in a tree, one section of which is shown in Fig. \ref{fig:tree model 1}. In other words, each $x_1[m]$ is a root with two children $x_2[\psi_j^k]$. 
Specifically, we stipulate that if $x_1[k] = 0$, then $x_2[m] = 0 \; \forall m \in \mathscr{T}^k$. 
%Suppose that we impose the following tree-structured sparsity profile: If a root node is $0$, then all descendant nodes must also be $0$.
%This means that if $x_1[1] = 0$, both $x_2[1]$ and $x_2[2]$ must also be $0$. 
Hierarchical sparsity was first studied in \cite{kim2010tree,jenatton2011structured,zhao2009composite} and later incorporated into a unimodal dictionary learning framework in \cite{jenatton2010proximal}. Later, Bayesian hierarchical sparse signal recovery techniques were developed, which form the basis for the following derivation \cite{zhang2014image,zhang2013fast}.

%More formally, tree sparsity can be restated as: If $\Vert x_1[\psi_1^k] \Vert_0 = 0$, then $\Vert x_2[\psi_2^k] \Vert_0 = 0$, \textit{but not vice versa} (as in Section \ref{sec:subspace2subspace}).
%More formally, let the tree describing the coefficient structure be spanned by $2 M_2$ groups consisting of every node in the tree along with its descendants. To keep track of which coefficients belong to each group, we define $g_j \in [M_j \cup 0]^{2 M_2}$, where $g_j[k] = m$ means that coefficient $m$ from modality $j$ is a member of group $k$. We use $g_j[k] = 0$ to denote that no coefficients from modality $j$ are members of group $k$. 
From an optimization point of view, hierarchical sparsity can be promoted through a composite regularizer \cite{zhao2009composite}. In this case, the regularizer could\footnote{The exact form of the regularizer depends on how the energy in a given group is measured.} take the form
\begin{align}\label{eq:tree regularizer}
\sum_{k \in [K], m\in \mathscr{T}^k}   \left\Vert \begin{bmatrix}
x_1[k] & x_2\left[ m \right]
\end{bmatrix} \right\Vert_2 +  \vert x_2[m] \vert.
\end{align}
As described in \cite{zhao2009composite}, the key to designing a composite regularizer for a given root-leaf pair is to measure the group norm of the pair along with the energy of the leaf alone. The combination of the group and individual norms serve two purposes which, jointly, promote hierarchical sparsity \cite{zhao2009composite}: 
\begin{enumerate}[label=\textbf{R\arabic*}]
\item It is possible that $x_2[m] = 0, m \in \T^k$, without requiring $x_1[k] = 0$.
\item \label{R:2} The infinitesimal penalty on $x_1[k]$ deviating from $0$ tends to $0$ for $\vert x_2[m] \vert > 0, m \in T^k$. 
\end{enumerate}
The choice of $\ell_2$-norm in \eqref{eq:tree regularizer} guarantees that the regularizer satisfies \ref{R:2} [Theorem 1, \cite{zhao2009composite}]. The conditions of [Theorem 1, \cite{zhao2009composite}] are violated if the $\ell_2$ norm is replaced by an $\ell_1$ norm in \eqref{eq:tree regularizer}. 

In a Bayesian setting, we can mimic the effect of the regularizer in \eqref{eq:tree regularizer} through an appropriately defined prior on $\bm{x}$. Let 
\begin{small}
\begin{align}\label{eq:overlapping}
\tilde{x}_j &= S_j x_j , \; \; S_1 \in \mathbb{B}^{M_2 \times M_1}, \; \; S_2 = \begin{bmatrix}
\mathsf{I} & \mathsf{I}
\end{bmatrix}^T \in \mathbb{B}^{2 M_2 \times M_2}
\end{align}
\end{small}
where $S_1$ is a binary matrix such that $S_1[m,k] = 1$ if and only if $m \in \mathscr{T}^k$. 
%The regularizer in \eqref{eq:tree regularizer} can then be re-written as
%\begin{align}
%\left\Vert \begin{bmatrix}
%\tilde{x}_1 & \tilde{x}_2
%\end{bmatrix} \right \Vert_{21}
%\end{align}
Let $R_j$ be a diagonal matrix such that $S_j^T S_j R_j = \mathsf{I}$\footnote{A diagonal $R_j$ is guaranteed to exist because $S_j^T S_j$ is itself a diagonal matrix.} and define $\hat{x}_j = S_j R_j x_j$. Let ${\gamma_j} \in \mathbb{R}^{M_2} \; \forall j$ and
\begin{align}\label{eq:tree prior}
\p{\bm{x} | \bm{\gamma}} &= \mathsf{N}\left(\hat{x}_1 ; 0, \Gamma_1 \right)  \mathsf{N}\left(\hat{x}_2 ; 0, \Gamma_{2} \right),
\end{align}
where $\Gamma_1 = diag(\gamma_1)$ and $\Gamma_{2} = diag\left(\begin{bmatrix}
\gamma_1^T & \gamma_2^T
\end{bmatrix}^T \right)$. A prototype branch for the hierarchical sparsity prior is shown in Fig. \ref{fig:hierarchical}. To see how this model leads to hierarchical sparsity, observe that
\begin{align}
\label{eq:hierarchical sparsity}
\begin{split}
\p{x_1[k] | \bm{\gamma}} &= \function{\mathsf{N}}{0,\gamma_1[k]}\\
\p{x_2[m] | \bm{\gamma}} &= \function{\mathsf{N}}{0,\inv{\gamma_1^{-1}[k]+\gamma_2^{-1}[m]}}
\end{split}
\end{align} 
for $m \in \mathscr{T}^k$. If we infer that $\gamma_1[k] = 0$, then the prior on both $x_1[k]$ and $x_2[m], \forall m \in \mathscr{T}^k$, reduces to a dirac-delta function, i.e. if the root is zero, then the leaves must also be zero. On the other hand, if $\gamma_2[m]$ is inferred to be $0$, only the prior on $\gamma_2[m]$ is affected, i.e. leaf sparsity does not imply root sparsity.

Inference for the tree-structured model proceeds in a similar fashion to that shown in Section \ref{sec:msbdl inference}, with a few variations. The goal is to optimize \eqref{eq:ML} through the EM algorithm. The difference here is that we use $ \braces{ \bm{\hat{X}}, \bm{Y}, \theta }$ and  $\bm{\hat{X}}$ as the complete and nuisance data, respectively. In order to carry out inference, we must first find the posterior density $\p{\bm{\hat{X}} | \bm{Y},\theta}$. It is helpful to first derive the signal model in terms of $\bm{\hat{x}}$ \cite{zhang2014image}:
\begin{align}\label{eq:tree signal model}
\p{y_j | D_j,\hat{x}_j,\sigma_j} &= \mathsf{N}\left( A_j S_j^T \hat{x}_j,\sigma_j^2 \mathsf{I} \right).
\end{align}
Using \eqref{eq:tree signal model}, it can be shown that
\begin{align}
\p{\hat{x}_j^i | y_j^i,\theta} &= \mathsf{N}\left(\mu_{\hat{x},j}^i,\Sigma_{\hat{x},j}^i \right)\\
\Sigma_{\hat{x},j}^i &= \inv{\sigma_j^{-2} S_j D_j^T D_j S_j^T + \inv{\Gamma_j^i}}\\
\mu_{\hat{x},j}^i &= \sigma_j^{-2} \Sigma_{\hat{x},j}^i D_j S_j^T y_j.
\end{align}
The likelihood function itself is different from \eqref{eq:log likelihood}-\eqref{eq:Sigma y} and is given by
$
\p{y_j^i | \theta} = \mathsf{N}\left(0,\Sigma_{\hat{y},j}^i \right)$, where $
\Sigma_{\hat{y},j}^i = \sigma_j^2 \mathsf{I} + D_j S_j^T \Gamma_j^i S_j D_j^T$.
The EM update rules are given by
\begin{footnotesize}
\begin{align}\label{eq:gamma update hierarchical}
\left(\gamma_j^i[m]\right)^{t+1} &= \begin{cases}
0.5 \sum_{j'=1}^2 \Sigma_{\hat{x},j'}^i[m,m]+\left(\mu_{\hat{x},j'}^i[m]\right)^2 & \text{ if } m \leq M_2 \\
\Sigma_{\hat{x},j}^i[m,m]+\left(\mu_{\hat{x},j}^i[m]\right)^2 & \text{ else. }
\end{cases} \\ 
\label{eq:D update hierarchical}
\left(D_j\right)^{t+1} &= Y_j U_j^T S_j\inv{ S_j^T \left( U_j U_j^T + \sum_{i \in [L]} \Sigma_{\hat{x},j}^i\right) S_j}.
\end{align}
\end{footnotesize}
Extending the hierarchical sparsity prior for $J > 2$ is straightforward and depends on the specific application being considered. It is possible to simply append more leaves to each root $x_1[k]$ corresponding to coefficients from modality $j > 2$. Another possibility is to treat each $x_2[m]$ as itself a root with leaves from $x_3$, assigning a hyperparameter to each $x_2$-$x_3$ root-leaf pair as well as to each $x_3$ leaf, and repeating the process until all modalities are incorporated into the tree.

%\subsection{Inverted hierarchical sparsity}
%\begin{figure}
%\centering
%\tikz{
%\node[latent] (x21) {$x_{2}[m_{2,1}]$};
%\node[above=of x21](dots){$\cdots$};
%\node[latent, left=of dots] (x11) {$x_1[m_{1,1}]$};
%\node[latent, right=of dots] (x12) {$x_1[m_{1,P_k}]$};
%\edge[-]{x21}{x11,x12};
%}
%\caption{}
%\label{fig:tree model 2}
%\end{figure}

\subsection{Avoiding Poor Stationary Points}
\label{sec:poor stationary points}
For both the atom-to-subspace and hierarchical sparsity models, we experimentally observe that MSBDL tends to get stuck in undesirable stationary points. In the following, we describe the behavior of MSBDL in these situations and offer a solution. Suppose that data is generated according to the atom-to-subspace model, where $D_j$ denotes the true dictionary for modality $j$. In this scenario, we experimentally observe that MSBDL performs well when $\vert \mathscr{T}^k \vert = c \; \forall k$. On the other hand, if $\vert \mathscr{T}^k \vert$ varies as a function of $k$, MSBDL tends to get stuck in poor stationary points, where the quality of a stationary point is (loosely) defined next. Let $\vert \mathscr{T}^k \vert = 1$ for all $k$ except $k'$, for which $\vert \mathscr{T}^{k'}\vert = 2$, i.e. $M_2 = M_1+1$. In this case, MSBDL is able to recover $D_1$, but recovers only the atoms of $D_2$ which are indexed by $\mathscr{T}^k \; \; \forall k \neq k'$. 
%We conjecture that MSBDL recovers the subspace/atoms of $D_2^*$ indexed by $\mathscr{T}^{k'}$ only if $d_1^{k'}$, the $k'$'th column of the estimated dictionary $D_1$, converges to ${d_1^*}^{k'}$.

To avoid poor stationary points, we adopt the following strategy. If the tree describing the assignment of columns of $D_2$ to those of $D_1$ is unbalanced, i.e. $\vert \mathscr{T}^k \vert$ varies with $k$, then we first balance the tree by adding additional leaves\footnote{A balanced tree is one which has the same number of leaves for each subtree, or $\vert \mathscr{T}^k \vert = c$.}. Let $\hat{M}_2$ be the number of leaves in the balanced tree. We run MSBDL until convergence to generate $\bm{\hat{D}}$. Finally, we prune away $\hat{M}_2 - M_2$ columns of $\hat{D}_2$ using the algorithm in Fig. \ref{alg:pruning}.

\begin{figure}
\centering
\begin{algorithmic}[1]
\Require $D,U,\lbrace \mathscr{T}^k \rbrace_{k \in [K]},M_p,S,\epsilon$
\State Let $z[m] = \norm{ \left(S^T U\right)[m,:] }_2^2 \; \forall m$ and
\begin{align*}
v[k] = \argmax_{m_1,m_2 \in \mathscr{T}^k \text{ s.t. } m_1 \neq m_2} \frac{ \vert \transpose{d^{m_1}} d^{m_2} \vert}{\Vert d^{m_1} \Vert_2 \Vert d^{m_2} \Vert_2}
\end{align*}
\While{$\exists k \text{ s.t. } v[k] > \epsilon \text{ and } \vert \mathscr{T}^k \vert > 1 \text{ and } M_p > 0$}
\State $k = \argmax_k v[k]$
\State Find $m = \argmin_{m \in \mathscr{T}^{k}} z[m]$
\State Remove column $m$ from $D$ and $\mathscr{T}^{k}$
\State $M_p \leftarrow M_p - 1$
\EndWhile

\While{$M_p > 0$}
\State Find
$
m = \argmin_{m \in \mathscr{T}^k \text{ s.t. } \vert \mathscr{T}^k \vert > 1} z[m]
$
\State Remove column $m$ from $D$ and $\mathscr{T}^{k}$
\State $M_p \leftarrow M_p - 1$
\EndWhile
\end{algorithmic}
\caption{Pruning algorithm for the learning strategy in Section \ref{sec:poor stationary points}. $M_p$ denotes the number of columns to be pruned, $S$ is given by the identity matrix for the atom-to-subspace model and by $S_2$ in \eqref{eq:overlapping} for the hierarchical sparsity model, and $U$ denotes the matrix of first order sufficient statistics.}
\label{alg:pruning}
\end{figure}

%It is insightful to consider why this issue does not occur for the dictionary model in \eqref{eq:gsm}. In that scenario, it is irrelevant whether $D_1$ converges to $D_1^*$ or $D_1^* H$, where $H$ is an arbitrary permutation matrix, as long as $D_2$ converges to $D_2^* H$. For the case 

\section{Task Driven MSBDL (TD-MSBDL)}
\label{sec:TD-MSBDL}
\begin{figure}
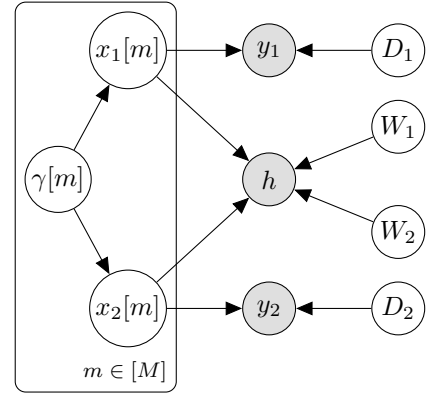

\centering
\tikz{
\node[latent] (x1) {$x_{1}[m]$};
\node[obs,right=of x1](y1){$y_{1}$};
\node[latent,right=of y1] (D1) {$D_{1}$};
\node[obs,below=of y1] (q) {$h$};

\node[latent,left=of q, xshift=-1cm](g1){$\gamma[m]$};
\node[obs,below= of q](y2){$y_{2}$};
\node[latent,right=of y2] (D2) {$D_{2}$};

\node[latent,left =of y2] (x2) {$x_{2}[m]$};

\node[latent,right=of q, yshift=0.7cm] (W1) {$W_{1}$};
\node[latent,right=of q, yshift=-0.7cm] (W2) {$W_{2}$};
\plate{}{(x1)(x2)(g1)}{$m \in [M]$};

\edge[->]{g1}{x1};
\edge[->]{g1}{x2};
\edge[->]{x1}{y1,q};
\edge[<-]{y1}{D1};
\edge[->]{x2}{y2,q};
\edge[<-]{y2}{D2};
\edge[<-]{q}{W1,W2};
}
\caption{Graphical model for two modality TD-MSBDL.}
\label{fig:graphical model 3}
\end{figure}

In the following, we describe a task-driven extension of the MSBDL algorithm. For purposes of exposition, we assume the one-to-one prior in \eqref{eq:gsm}, but the approach applies equally to the priors discussed in Section \ref{sec:modeling complex relationships}, as discussed in Section \ref{sec:results}. To incorporate task-driven learning, we modify the MSBDL graphical model to the one shown in Fig. \ref{fig:graphical model 3}. We set 
$
\p{ h | x_j, W_j} = \function{\mathsf{N}}{W_j x_j,\beta_j^2 \mathsf{I}}
$,
where $\beta_j$ is the class label noise standard deviation for modality $j$. The class label noise standard deviation is modality dependent, affording the model an extra level of flexibility compared to \cite{bahrampour,jiang2013label,mairal2012task}. The choice of the Gaussian distribution for the conditional density of ${h}$, as opposed to a multinomial or softmax, stems from the fact that the posterior $\p{x_j | y_j,h,{D}_j,{W}_j}$ needed to perform EM remains computable in closed form, i.e.
$
\p{x_j^i | y_j^i,h^i,{D}_j,{W}_j} = \mathsf{N}\left({\Sigma}_{x,j}^{TD,i} , {\mu}_{j}^{TD,i} \right)
$
where 
\begin{align}
{\Sigma}_{x,j}^{TD,i} &= \inv{\sigma_j^{-2} {D_j^T D_j} + \beta_j^{-2} W_j^T W_j + \inv{\Gamma^i}}\label{eq:Sigma update TD}\\
{\mu}_j^{TD,i} &= {\Sigma}_{x,j}^{TD,i} \left(\sigma_j^{-2} {D}_j^T y_j^i + \beta_j^{-2} {W}_j^T h^i \right) \label{eq:mu update TD}
\end{align}

\vspace{-1em}
\subsection{Inference Procedure}
We employ EM to optimize
\begin{align}\label{eq:TD log likelihood}
\argmax_{\theta^{TD}} \log \p{\bm{Y},H | \theta^{TD}}
\end{align}
where $\theta^{TD} = \left\lbrace \theta,\bm{W} \right \rbrace$. It can be shown that
$
\p{y_j,h | x_j,\theta^{TD}} = \mathsf{N}\left(\begin{bmatrix} y_j^T & h^T \end{bmatrix}^T ; 0,\Sigma_{y,j}^{TD} \right)
$
where
\begin{align}
\Sigma_{y,j}^{TD} = \begin{bmatrix}
\sigma_j^2 \mathsf{I} + D_j \Gamma D_j^T & 0 \\ 0 & \beta_j \mathsf{I} + W_j \Gamma W_j^T
\end{bmatrix}.
\end{align}
The update rules for $\bm{D}$ and $\ibraces{\gamma^i}$ \footnote{Assuming that the prior in \eqref{eq:gsm} is used.} remain identical to \eqref{eq:gamma update block} and \eqref{eq:D update}, respectively, with the exception that the modified posterior statistics shown in \eqref{eq:Sigma update TD}-\eqref{eq:mu update TD} are used. The update of $W_j$ is given by
\begin{align}
W_j^{t+1} &=  H \transpose{{U}_{j}^{TD}} \inv{{{U}_{j}^{TD}} \transpose{{U}_{j}^{TD}}+\sum_{i \in [L]} {\Sigma}_{x,j}^{TD,i}}\label{eq:W update old}\\
U_j^{TD} &= \begin{bmatrix}
\mu_j^{TD,1} & \cdots & \mu_j^{TD,L}
\end{bmatrix}.
\end{align}
We also find it useful to add regularization in the form of $\nu \sum_{j \in [J]} \Vert W_j \Vert_F^2$, leading to the update rule
\begin{scriptsize}
\begin{align}
W_j^{t+1} &=  \begin{bmatrix}
H \transpose{{U}_{j}^{TD}} & 0
\end{bmatrix} \inv{
\begin{bmatrix}
U_j^{TD} \transpose{U_j^{TD}}+\sum_{i \in [L]} \Sigma_{x,j}^{TD,i} & \sqrt{\nu} \mathsf{I}
\end{bmatrix}
}.\label{eq:W update}
\end{align}
\end{scriptsize}
TD-MSBDL has the same worst-case computational complexity as MSBDL with the benefit of supervised learning. 

\subsection{Complete Algorithm}
Supervised learning algorithms are ultimately measured by their performance on test data. While a given algorithm may perform well on training data, it may generalize poorly to test data \footnote{In the supervised learning community, lack of generalization to test data is commonly referred to as over-fitting the training data.}. To maintain the generalization properties of the model, it is common to split the training data into a training set $\braces{\bm{Y},H}$ and validation set $\braces{\bm{Y^{V}},H^{V}}$, where the number of training points $L$ does not necessarily have to equal the number of validation points $L^{V}$. The validation set is then used during the training process as an indicator of generalization, as summarized by the following rule of thumb: Continue optimizing $\theta^{TD}$ until performance on the validation set stops improving. In the context of TD-MSBDL, the concept of generalization has a natural Bayesian definition: The parameter set $\braces{ \theta^{TD},\bm{\beta} }$ which achieves optimal generalization solves
\begin{align}
\argmax_{\theta^{TD} \in \mathscr{H} ,\bm{\beta}} \prod_{j \in [J]} \p{H^V | \theta^{TD},\beta_j},
\end{align}
where $\mathscr{H}$ denotes the set of solutions to \eqref{eq:TD log likelihood}. Note that $\p{H^V | \theta^{TD},\beta_j} = \p{H^V | W_j , \beta_j}$, which is intractable to compute since it requires integrating
$
\p{h^{V,i} | W_j , \gamma^{V,i}, \beta} = \mathsf{N}\left(0, \beta_j \mathsf{I} + W_j \Gamma^{V,i} W_j^T\right)
$
over $\gamma^{V,i}$, where $\Gamma^{V,i} = diag(\gamma^{V,i})$. As such, we approximate $ \p{H^V | W_j , \beta_j}$ by $ \p{H^V | W_j , \gamma^{*,V,i},\beta_j}$, where $\gamma^{*,V,i}$ is the output of MSBDL with fixed $D_j$ for input data $y_j^{V,i}$, leading to the tractable optimization problem
\begin{align}\label{eq:tractable generalization}
\argmax_{\theta^{TD} \in \mathscr{H} ,\bm{\beta}} \prod_{j \in [J]} \p{H^V | W_j , \braces{ \gamma^{*,V,i} }_{i \in [L^V]},  \beta_j}.
\end{align}
What remains is to select $\bm{\beta}$. As $\beta_j$ decreases, TD-MSBDL fits the parameters $\theta^{TD}$ to the training data to a larger degree, i.e. the optimizers of \eqref{eq:TD log likelihood} achieve increasing objective function values. Since direct optimization of \eqref{eq:tractable generalization} over $\bm{\beta}$ presents the same challenges as the optimization of $\bm{\sigma}$, we propose an annealing strategy which proposes progressively smaller values of $\beta_j$ until the objective in \eqref{eq:tractable generalization} stops improving:

\begin{footnotesize}
\begin{align}\label{eq:beta update}
\beta_j^{t+1} = 
\begin{cases}
\tilde{\beta}_j^{t+1} & \text{ if } \log \p{H^V | W_j , \braces{ \gamma^{*,V,i} }_{i \in [L^V]}, \tilde{\beta}_j^{t+1}} > \\ & \log \p{H^V | W_j , \braces{ \gamma^{*,V,i} }_{i \in [L^V]}, {\beta}_j^t}\\
\beta_j^t & \text{ else }
\end{cases}
\end{align}
\end{footnotesize}%
where $\beta_j^0 > \beta^\infty \geq 0,\alpha_\beta < 1, \tilde{\beta}_j^{t+1} = \max(\beta^\infty,\alpha_\beta \beta_j^t)$, and we make the same recommendations for setting $\bm{\beta^0}$ and $\beta^\infty$ as $\bm{\sigma^0}$ and $\sigma^\infty$ in Section \ref{sec:complete msbdl algorithm}. Computing \eqref{eq:beta update} is computationally intensive because it requires running MSBDL, so we only update $\log \p{H^V | W_j , \braces{ \gamma^{*,	V,i} }_{i \in [L^V]}, {\beta}_j^t}$ every $T^V$ iterations. Due to space considerations, the complete TD-MSBDL algorithm is provided in the Supplementary Material. Given test data $Y_j^{test}$, we first run MSBDL with $D_j$ fixed and treat $\mu_j^{test,i}$ as an estimate of $x_j^{test,i}$. The data is then classified according to $
\argmax_{c \in [C]} \left( W_j \mu_j^{test,i} \right)[c]
$,
where $e_c$ refers to the $c$'th standard basis vector \cite{bahrampour}.

\section{Analysis}
%\subsection{Annealing}
\label{sec:analysis}
We begin by analyzing the convergence properties of MSBDL. Note that MSBDL is essentially a block-coordinate descent algorithm with blocks $\theta$ and $\bm{\sigma}$. Therefore, we first analyze the $\theta$ update block, then the $\bm{\sigma}$ update block, and finally the complete algorithm. Unless otherwise specified, we assume a non-informative prior on $\gamma$ and the one-to-one prior\footnote{Extension of Theorems \ref{thm:stationary D gamma}-\ref{thm:convergence of MSBDL} to the atom-to-subspace and hierarchical priors is straightforward, but omitted for brevity.}. While MSBDL relies heavily on EM, it is not strictly an EM algorithm because of the $\bm{\sigma}$ annealing procedure. It will be shown that MSBDL still admits a convergence guarantee and an argument will be presented for why annealing $\bm{\sigma}$ produces favorable results in practice. Although we focus specifically on MSBDL, the results can be readily extended to TD-MSBDL, with details omitted for brevity. Proofs for all results are shown in the Supplementary Material. 

We first prove that the set of iterates $\braces{\theta^t}_{t=1}^\infty$ produced by the inner loop of MSBDL converge to the set of stationary points of the log-likelihood with respect to $\theta$. This is established by proving that the objective function is coercive (Theorem \ref{thm:coercive}), which means that $\braces{\theta^t}_{t=1}^\infty$ admits at least one limit point (Corollary \ref{cor:existence of limit point1}), and then proving that limit points are stationary (Theorem \ref{thm:stationary D gamma}).
\begin{theorem}
\label{thm:coercive}
Let $\jbraces{\sigma_j > 0}$, let there be at least one $j^*$ for each $m$ such that $\norm{ d_{j^*}^m}_2 > 0$, and let $\exists i$ such that $\gamma^i[m] > 0$ for any choice of $m$. Then, $-\log \p{\bm{Y} | \theta,\bm{\sigma}}$ is a coercive function of $\braces{ \theta,\bm{\sigma}}$.
\end{theorem}

\begin{corollary}
\label{cor:existence of limit point1}
Let the conditions of Theorem \ref{thm:coercive} be satisfied. Then, the sequence $\braces{\theta^t }_{t=1}^\infty$ produced by the inner loop of the MSBDL algorithm admits at least one limit point.
\end{corollary}

The proof of Corollary \ref{cor:existence of limit point1} is shown in the Appendix, but the high level argument is that $\braces{\theta^t }_{t=1}^\infty$ is a member of a compact set. As such $\braces{\theta^t }_{t=1}^\infty$ admits at least one convergent subsequence. Since there may be multiple convergent subsequences, Corollary \ref{cor:existence of limit point1} does not preclude $\braces{\theta^t }_{t=1}^\infty$ having multiple limit points, or a limit set. The point is that $\braces{\theta^t }_{t=1}^\infty$ admits one or more limit points, and Theorem \ref{thm:stationary D gamma} below proves each of these points to be stationary.
\begin{theorem}
\label{thm:stationary D gamma}
Let the conditions of Corollary \ref{cor:existence of limit point1} be satisfied, $D_j^t$ be full rank for all $t$ and $j$, $\bm{\sigma}$ be fixed, and generate $\lbrace \theta^t\rbrace_{t=1}^\infty$ using the inner loop of MSBDL. Then, $\lbrace \theta^t \rbrace_{t=1}^\infty$ converges to the set of stationary points of $\log \p{\bm{Y} | \theta, \bm{\sigma}}$. Moreover, $\left\lbrace \log \p{\bm{Y} | \theta, \bm{\sigma}}\right\rbrace_{t=1}^\infty$ converges monotonically to $\log \p{\bm{Y} | \theta^*, \bm{\sigma}}$, for stationary point $\theta^*$.
\end{theorem}
The requirement that $D_j^t$ be full rank is easily satisfied in practice as long as $L > N_j$. Note that the SBL algorithm in \cite{wipf2004sparse} is a special case of MSBDL for fixed $\bm{D}$ and $J=1$. To the best of our knowledge, Theorem \ref{thm:stationary D gamma} represents the first result in the literature guaranteeing the convergence of SBL. A similar result to Theorem \ref{thm:stationary D gamma} can be given in the stochastic EM regime. 
\begin{theorem}
\label{thm:unique M step}
Let $\bm{\sigma}$ be fixed, $U_j^t$ be full rank for all $t,j$, and generate $\lbrace {\theta}^t \rbrace_{t=1}^\infty$ using the inner loop of MSBDL, only updating the sufficient statistics for a batch of points at each iteration (i.e. incremental EM). Then, the limit points of $\lbrace {\theta}^t \rbrace_{t=1}^\infty$ are stationary points of $\log \p{\bm{Y} | \theta, \bm{\sigma} }$.
\end{theorem}
If we consider the entire MSBDL algorithm, i.e. including the update of $\bm{\sigma}$, we can still show that MSBDL is convergent.
\begin{corollary}
\label{cor:existence of limit point2}
Let the conditions of Theorems \ref{thm:coercive} and \ref{thm:stationary D gamma} be satisfied. Then, the sequence $\braces{\theta^t,\bm{\sigma^t} }_{t=1}^\infty$ produced by the MSBDL algorithm admits at least one limit point.
\end{corollary}

Although the preceding results establish that MSBDL has favorable convergence properties, the question still remains as to why we choose to anneal $\sigma_j$. At a high level, it can be argued that setting $\sigma_j$ to a large value initially and then gradually decreasing it prevents MSBDL from getting stuck in poor stationary points with respect to $\theta$ at the beginning of the learning process. To motivate this intuition, consider the log likelihood function in \eqref{eq:ML}, which decomposes into a sum of $J$ modality-specific log-likelihoods. The curvature of the log-likelihood for modality $j$ depends directly on $\sigma_j$. Setting a high $\sigma_j$ corresponds to choosing a relatively flat log likelihood surface, which, from an intuitive point of view, has less stationary points. This intuition can be formalized in the scenario where $D_j$ is constrained in a special way.
\begin{theorem}\label{thm:number of local minima}
Let $\sigma_j^1 > \sigma_j^2, \Psi_j = \lbrace D_j: D_j = \begin{bmatrix}
\check{D}_j & \mathsf{I}
\end{bmatrix} \rbrace$, and $\Omega_{\sigma_j,j} = \left\lbrace \Sigma_{y,j} : \Sigma_{y,j} = \sigma_j^2 \mathsf{I} + D_j \Gamma {D_j}^T, D_j \in \Psi_j \right\rbrace$. Then, $\Omega_{\sigma_j^1,j} \subseteq \Omega_{\sigma_j^2,j} $.
\end{theorem}
Theorem \ref{thm:number of local minima} suggests that as $\sigma_j$ gets smaller, the space over which the log-likelihood in \eqref{eq:ML} is optimized grows. As the optimization space grows, the number of stationary points grows as well\footnote{This holds only for the constrained scenario in Theorem \ref{thm:number of local minima}.}. As a result, it may be advantageous to slowly anneal $\sigma_j$ in order to allow MSBDL to learn $D_j$ without getting stuck in a poor stationary point.

If we constrain the space over which $D_j$ is optimized to $\Psi_j$, as in Theorem \ref{thm:number of local minima}, then we can establish a number of interesting convergence results.
\begin{theorem}\label{thm:convergence of sigma}
Let $\alpha_\sigma$ be arbitrarily close to $1$, $\sigma^\infty = 0$, $\sigma_j^0 \geq \argmax_{\sigma_j} \max_{\theta} \log \p{Y_j | \theta,\sigma_j}$, $D_j \in \Psi_j$, $\theta^t = \argmax_\theta \log \p{Y_j | \theta,\sigma_j^{t-1}}$, and consider updating $\sigma_j$ using \eqref{eq:sigma update}. Then, $\sigma_j^t = \sigma_j^{t-1}$ implies $\sigma_j^t = \argmax_{\sigma_j} \log \p{Y_j | \theta^{t},\sigma_j}$.
\end{theorem}
Theorem \ref{thm:convergence of sigma} states that, under certain conditions, annealing terminates for a given modality $j$ at a global maximum of the log-likelihood with respect to $\sigma_j$ for fixed $\theta^{t}$. The conditions of Theorem \ref{thm:convergence of sigma} ensure that \eqref{eq:sigma update} only terminates at stationary points of the log-likelihood.

\begin{theorem}\label{thm:convergence of MSBDL}
Let the conditions of Theorems \ref{thm:stationary D gamma} and \ref{thm:convergence of sigma} and Corollary \ref{cor:existence of limit point2} be satisfied. Then, the sequence $\left\lbrace {\theta}^t,\bm{\sigma^t} \right\rbrace_{t=1}^\infty$ produced by MSBDL converges to the set of stationary points of $\log \p{\bm{Y} | \theta,\bm{\sigma}}$. 
\end{theorem}
Convergence results like Theorem \ref{thm:convergence of MSBDL} cannot be established for K-SVD and J$\ell_0$DL because they rely on greedy search techniques. In addition, no convergence results are presented in \cite{bahrampour} for J$\ell_1$DL.

%Although the inner loop of MSBDL converges to the set of stationary points in the incremental EM regime, Theorem \ref{thm:convergence of MSBDL} cannot be extended to this scenario because the log-likelihood function is not necessarily monotonic for incremental EM \cite{gunawardana2005convergence}.

%\begin{theorem}
%Theorem stating that MSBDL cannot recover $D_2$ is $D_1$ is not perfectly recovered for atom-to-subspace/hierarchical models.
%\end{theorem}
%
%\begin{theorem}
%Theorem stating that pruning strategy is optimal.
%\end{theorem}

%\subsection{Guarantees for multimodal dictionary recovery}
Finally, we consider what guarantees can be given for dictionary recovery in the noiseless setting, i.e. $
Y_j = D_j X_j \; \; \forall j.
$
We assume $M_j = M\; \forall j$ and that $\bm{x^i}$ share a common sparsity profile for all $i$. We do not claim that the following result applies to more general cases when $M_j \neq M \; \forall j$ or different priors. The question we seek to answer is: Under what conditions is the factorization $Y_j = D_j X_j$ unique? Due to the nature of dictionary learning, uniqueness can only be considered up to permutation and scale. Two dictionaries $D^1$ and $D^2$ are considered equivalent if $D^1 = D^2 Z$, where $Z$ is a permutation and scaling matrix. Guarantees of uniqueness in the unimodal setting were first studied in \cite{aharon2006uniqueness}. The results relied on several assumptions about the data generation process.
\begin{assumption}
\label{ass:1}
Let $s = \Vert x_j^i \Vert_0 \; \forall i,j$, and $s < \frac{\text{spark}(D_j)}{2}$, where $\text{spark}(D_j)$ is the minimum number of columns of $D_j$ which are linearly dependent \cite{donoho2003optimally}. Each $x_j^i$ has exactly $s$ non-zeros.
\end{assumption}

\begin{assumption}\label{ass:2}
$Y_j$ contains at least $s+1$ different points generated from every combination of $s$ atoms of $D_j$.
\end{assumption}

\begin{assumption}
\label{ass:3}
The rank of every group of $s+1$ points generated by the same $s$ atoms of $D_j$ is exactly $s$. The rank of every group of $s+1$ points generated by different atoms of $D_j$ is exactly $s+1$.
\end{assumption}
\begin{lemma}[Theorem 3, \cite{aharon2006uniqueness}]
\label{lemma:aharon uniqueness}
Let Assumptions \ref{ass:1}-\ref{ass:3} be true. Then, $Y_j$ admits a unique factorization $D_j X_j$. The minimum number of samples required to guarantee uniqueness is given by $(s+1) \binom{M}{s}$.
\end{lemma}
Treating the multimodal dictionary learning problem as $J$ independent unimodal dictionary learning problems, the following result follows from Lemma \ref{lemma:aharon uniqueness}.
\begin{corollary}\label{cor:1}
Let Assumptions \ref{ass:1}-\ref{ass:3} be true for all $j$. Then, $Y_j$ admits a unique factorization $D_j X_j$ for all $j$. The minimum number of samples required to guarantee uniqueness is given by $J (s+1) \binom{M}{s}$.
\end{corollary}

As the experiments in Section \ref{sec:results} will show, there are benefits to jointly learning multimodal dictionaries. It is therefore interesting to inquire whether or not there are \textit{provable} benefits to the multimodal dictionary learning problem, at least from the perspective of the uniqueness of factorizations. To formalize this intuition, consider the scenario where some data points $i$ do not have data available for all modalities. Let the Boolean matrix $P \in \mathbb{B}^{J \times L}$ be defined such that $P[j,i]$ is $1$ if data for modality $j$ is available for instance $i$ and $0$ else. The conditions on the amount of data needed to guarantee unique recovery of $\bm{D}$ by Corollary \ref{cor:1} can be restated as $L = (s+1) \binom{M}{s}$ and $P[j,i] = 1 \; \forall j,i$. The natural question to ask next is: Can uniqueness of factorization be guaranteed if $P[j,i] = 0$ for some $(j,i)$?
\begin{theorem}
\label{thm:uniqueness of multimodal DL}
Let $\bm{x^i}$ share a common sparsity profile for all $i$ and Assumptions \ref{ass:1}-\ref{ass:2} be true for all $j$. Let Assumption \ref{ass:3} be true for a single $j^*$. Let $G_{j}^k = \braces{ i: P[j,i] = 1 \text{ and } y_{j}^i \in \function{\text{span}}{D_j[:,\Upsilon^k]}}$ where $\Upsilon^k$ is the $k$'th subset of size $s$ of $[M]$. Let $\vert G_j^k \cap G_{j^*}^k \vert \geq s$ for all $j \neq j^*$ and $k$. Then, the factorization $Y_j = D_j X_j$ is unique for all $j$. The minimum total number of data points required to guarantee uniqueness is given by $J\left(s+\frac{1}{J}\right)\binom{M}{s}$.
\end{theorem}
Theorem \ref{thm:uniqueness of multimodal DL} establishes that the number of samples required to guarantee a unique solution to the multimodal dictionary learning problem is \textit{strictly less} than in Corollary \ref{cor:1}.

\section{Results}
\label{sec:results}
\subsection{Synthetic Data Dictionary Learning}
\label{sec:synthetic data results}
To validate how well MSBDL is able to learn unimodal and multimodal dictionaries, we conducted a series of experiments on synthetic data. We adopt the setup from \cite{yang2016sparse} and generate ground-truth dictionaries $D_j \in \mathbb{R}^{20\times 50}$ by sampling each element from a $\mathsf{N}(0,1)$ distribution and scaling the resulting matrices to have unit $\ell_2$ column norm. We then generate $x_j^i$ by randomly selecting $s=5$ indices and generating the non-zero entries by drawing samples from a $\mathsf{N}(0,1)$ distribution. The supports of $\bm{x^i}$ are constrained to be the same, while the coefficients are not. The elements of $v_j^i$ are generated by drawing samples from a $\mathsf{N}(0,1)$ distribution and scaling the resulting vector in order to achieve a specified Signal-to-Noise Ratio (SNR). We use $L = 1000$ and simulate both bimodal and trimodal datasets. The bimodal dataset consists of $30$dB ($j=1$) and $10$dB ($j=2$) SNR modalities. The trimodal dataset consists of $30$ dB ($j=1$), $20$ dB ($j=2$), and $10$ dB ($j=3$) SNR modalities. We use the empirical probability of recovering $\bm{D}$ as the measure of success, which is given by 
\begin{align}
&\frac{1}{M_j}\sum_{m \in [M_j]} \mathbbm{1}\left[\iota \left(d_j^m,\hat{D}_j\right) > 0.99\right]\\
&\iota \left(d_j^m,\hat{D}_j\right) = \max_{1 \leq m' \leq M_j} \frac{ \left \vert \transpose{d_j^m} \hat{d}_j^{m'} \right \vert}{\norm{d_j^m}_2 \norm{d_j^{m'}}_2},
\end{align}
where $\hat{D}_j$ denotes the output of the dictionary learning algorithm
and $\mathbbm{1}\left[ \cdot \right]$ denotes the indicator function. The experiment is performed $50$ times and averaged results are reported. We compare MSBDL with $\ell_1$DL\footnote{\url{http://spams-devel.gforge.inria.fr/downloads.html}}, K-SVD\footnote{\url{http://www.cs.technion.ac.il/~ronrubin/software.html}}, J$\ell_0$DL, and J$\ell_1$DL. While code for J$\ell_1$DL was publicly available, it could not be run on any of our Windows or Linux machines, so we used our own implementation. Code for J$\ell_0$DL was not publicly available, so we used our own implementation. For all algorithms, the batch size was set to $L$.

\begin{figure}
\centering
\begin{subfigure}[t]{0.45\columnwidth}
\centering
\includegraphics[width=\linewidth]{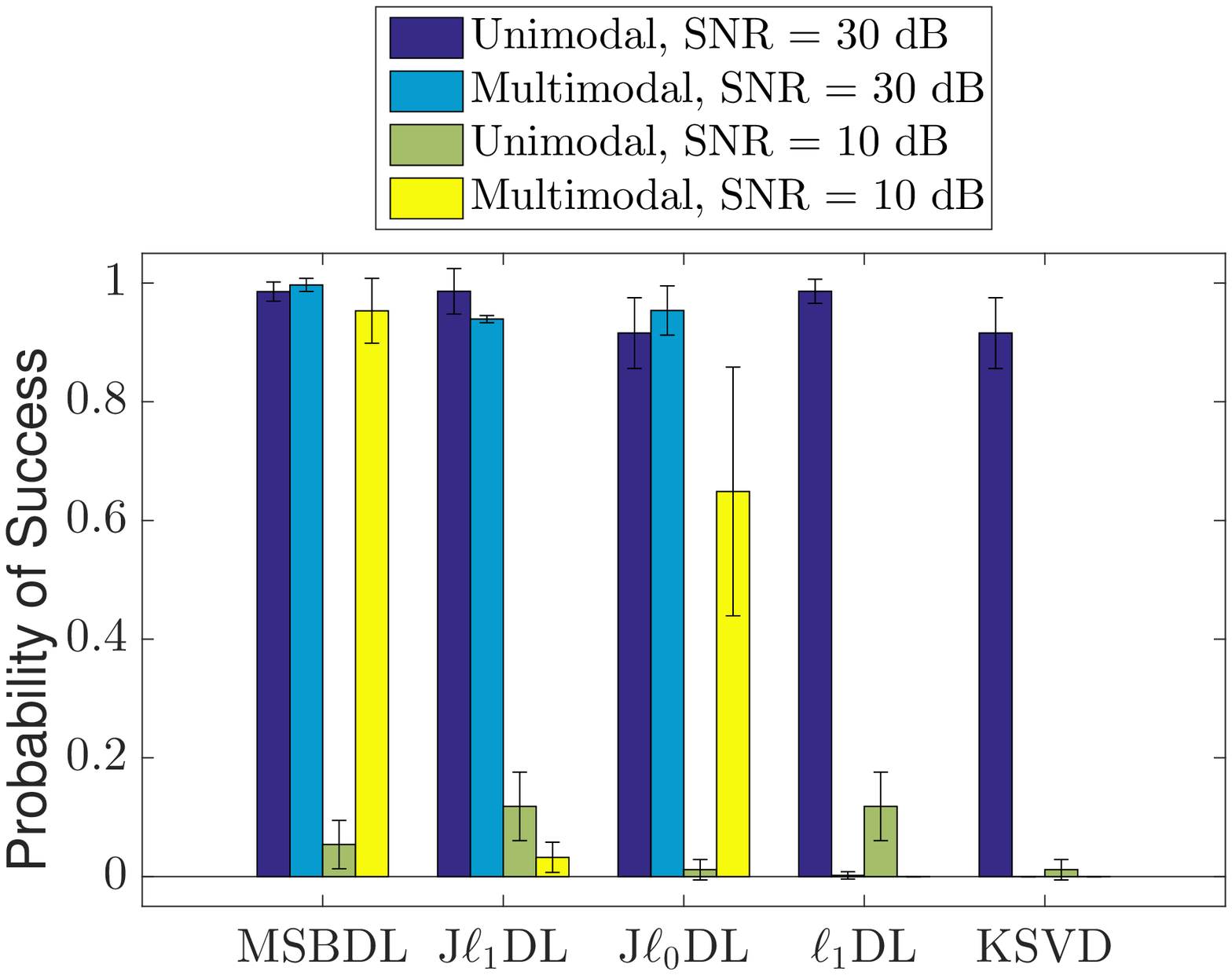}
\caption{}
\label{fig:bimodal synthetic}
\end{subfigure}
~
\begin{subfigure}[t]{0.45\columnwidth}
\centering
\includegraphics[width=\linewidth]{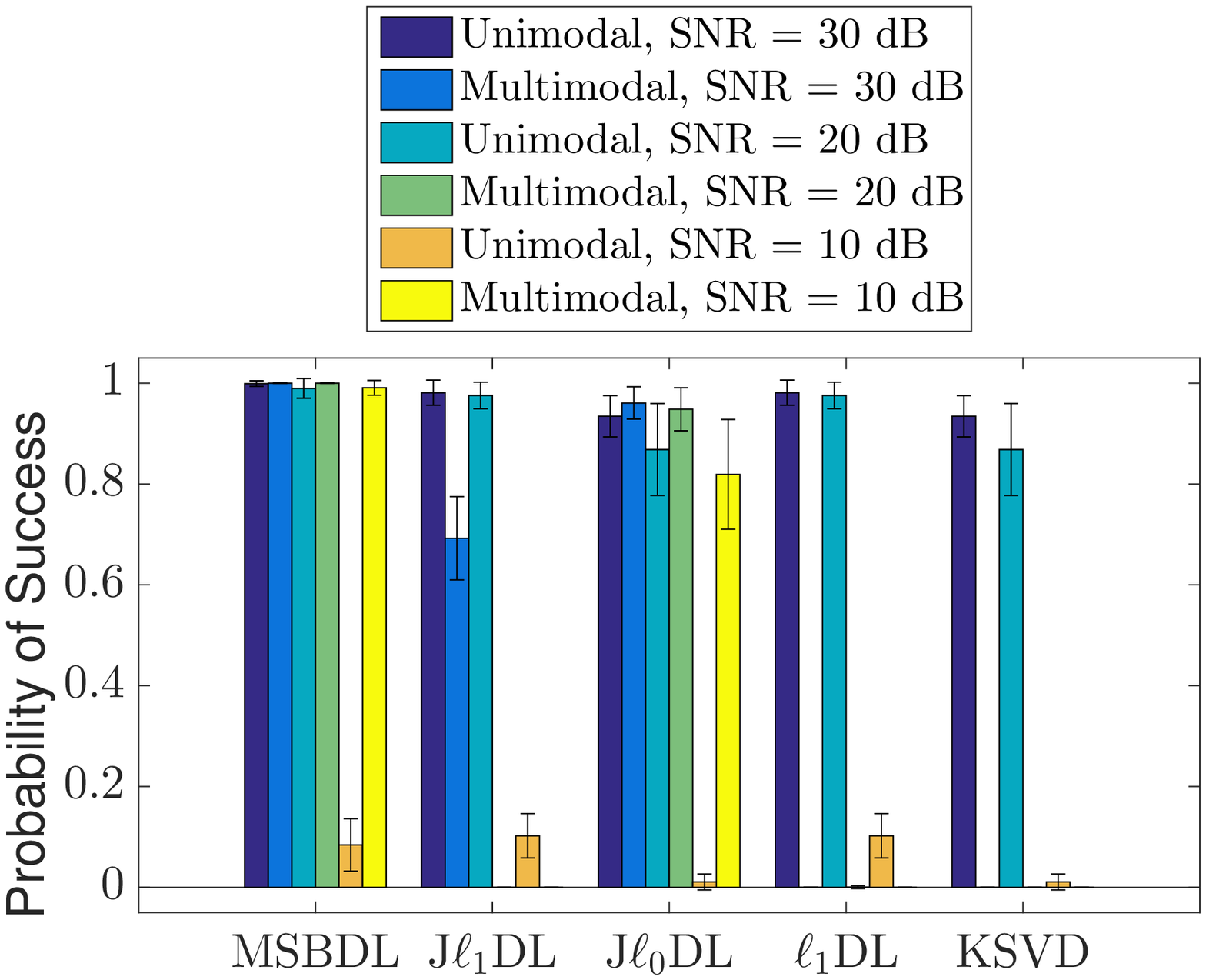}
\caption{}
\label{fig:trimodal synthetic}
\end{subfigure}
\caption{Bimodal (\ref{fig:bimodal synthetic}) and trimodal (\ref{fig:trimodal synthetic}) synthetic data results with one standard deviation error bars.}
\end{figure}

\begin{figure}
\centering
\begin{subfigure}{0.45\columnwidth}
\centering
\includegraphics[width=\linewidth]{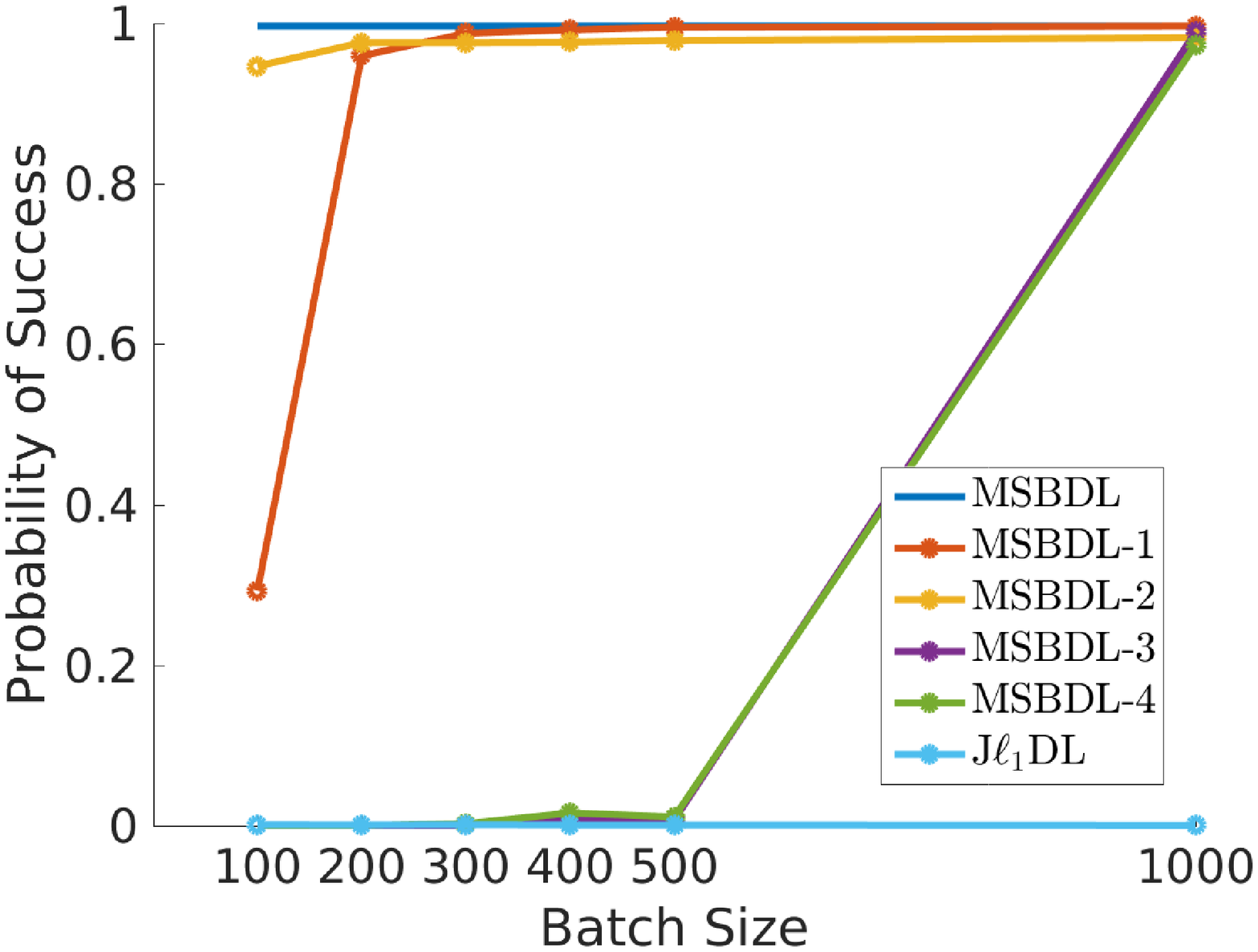}
\caption{}
\label{fig:batch_30db}
\end{subfigure}
~
\begin{subfigure}{0.45\columnwidth}
\centering
\includegraphics[width=\linewidth]{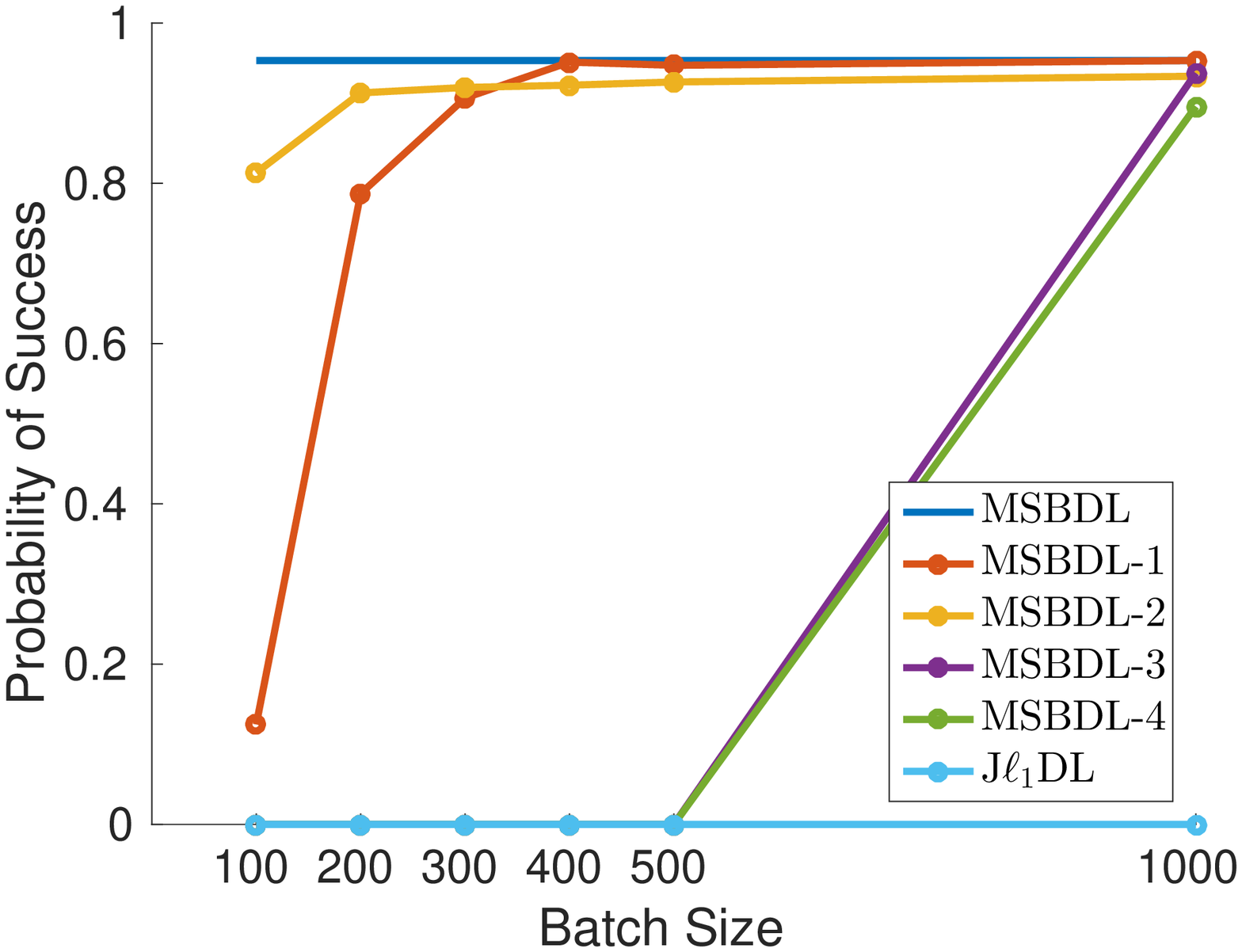}
\caption{}
\label{fig:batch_10db}
\end{subfigure}
\caption{Bimodal synthetic data results using stochastic learning for $30$ dB (Fig. \ref{fig:batch_30db}) and $10$ dB (Fig. \ref{fig:batch_10db}) datasets.}
\label{fig:stochastic}
\end{figure}

For the bimodal setting, the parameters $\sigma_1^0$ and $\sigma_2^0$ were set to $1$ and $\sqrt{10}$, respectively. For the trimodal setting, the parameters $\sigma_1^0,\sigma_2^0,$ and $\sigma_3^0$ were set to $1,\sqrt{1.5},$ and $\sqrt{2}$, respectively. In both cases, we set $\alpha_\sigma = \sqrt{0.995}$ and $\sigma^\infty = \sqrt{1e-3}$, where this choice of $\sigma^\infty$ corresponds to the lowest candidate $\lambda$ in the cross-validation procedure for competing algorithms. It was experimentally determined that MSBDL is relatively insensitive to the choice of $\sigma_j^0$ as long as $\sigma_1^0 < \sigma_2^0 < \sigma_3^0$, thus obviating the need to cross-validate these parameters. The regularization parameters ${\lambda}$ in \eqref{eq:l1} and $\bm{\lambda}$ in \eqref{eq:j0dl} were selected by a grid search over $\lbrace 1e-3,1e-2,1e-1,1\rbrace$ and both K-SVD and J0DL were given the true $s$. The parameter $\lambda_1$ was set to $1$ for J$\ell_0$DL across all experiments because the objective function in \eqref{eq:j0dl} depends only on the relative weighting of modalities. 
%For multimodal dictionary learning, K-SVD was given data in the format $\tilde{Y}$, like in \eqref{eq:l1}. 
All algorithms were run until convergence. 

The bimodal and trimodal dictionary recovery results are shown in Fig.'s \ref{fig:bimodal synthetic} and \ref{fig:trimodal synthetic}, respectively. For unimodal data, all of the algorithms recover the true dictionary almost perfectly when the $\text{SNR }= 30 \text{ dB}$, with the exception of J0DL and K-SVD. All of the tested algorithms perform relatively well for data with $20$ dB SNR and poorly on data with $10$ dB SNR, although MSBDL outperforms the other tested method in these scenarios. In the multimodal scenario, the proposed method clearly distinguishes itself from the other methods tested. For trimodal data, not only does MSBDL achieve $100\%$ accuracy on the $30$ dB data dictionary, but it achieves accuracies of $100\%$ and $99.2\%$ on the $20$ dB and $10$ dB data dictionaries, respectively. MSBDL outperforms the next best method by $17.2\%$ on the $10$ dB data recovery task\footnote{Throughout this work, we report the improvement to the probability of success or the classification rate in absolute terms.}. J$\ell_0$DL was able to capture some of the multimodal information in learning the $10$ dB data dictionary, but the $10$ dB data dictionary accuracy only reaches $81.9\%$. J$\ell_1$DL performs even worse in recovering the $10$ dB data dictionary, achieving $0\%$ accuracy. Similar trends can be seen in the bimodal results.

\begin{figure}
    \centering
    \begin{minipage}{0.45\columnwidth}
        \centering
        \includegraphics[width=\linewidth]{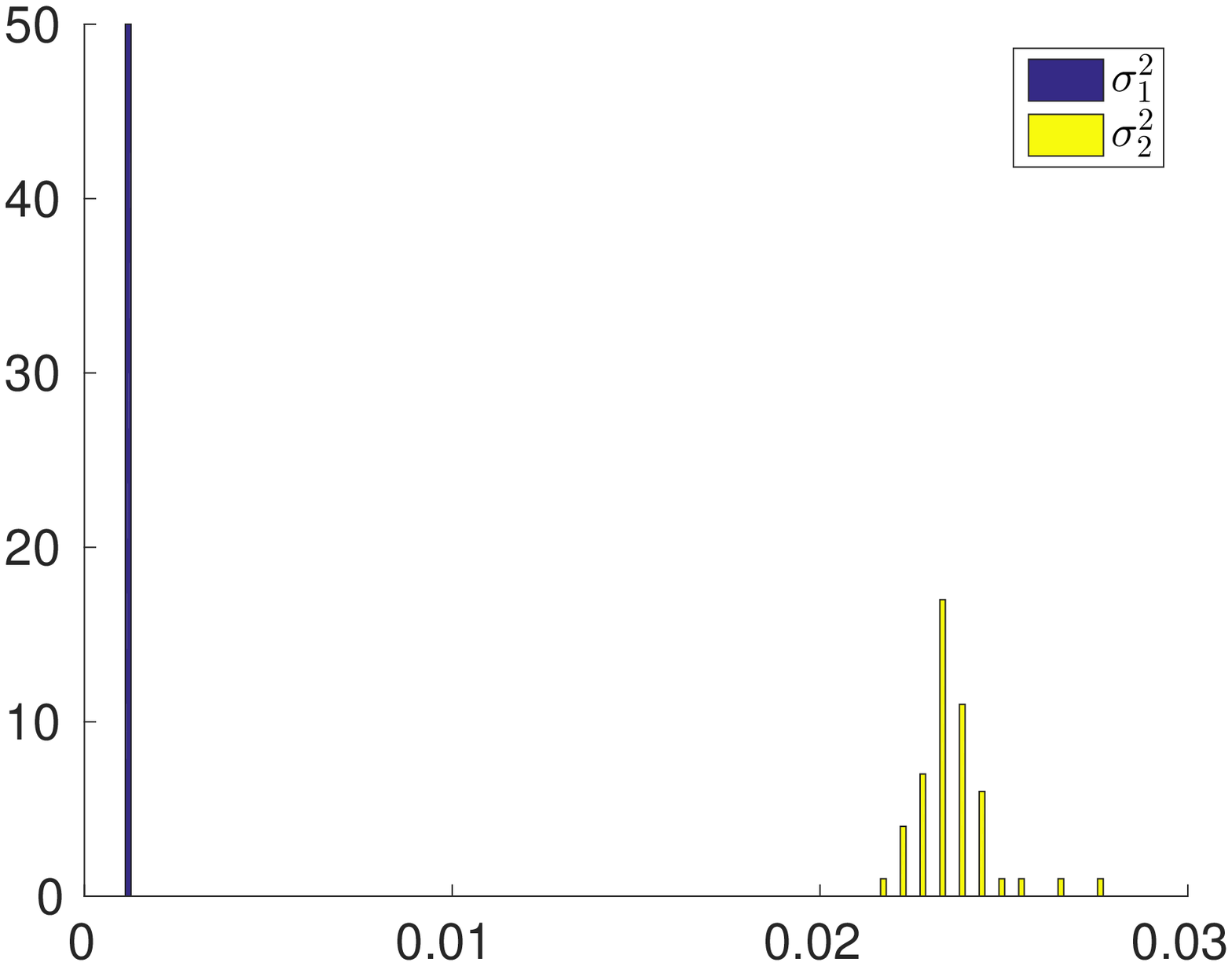}
\caption{Histogram of $\sigma_j^2$ at convergence for a bimodal dataset consisting of $30$ dB and $10$ dB modalities.}
\label{fig:hist_s2}
    \end{minipage}\hfill
    \begin{minipage}{0.45\columnwidth}
        \centering
        \includegraphics[width=\linewidth]{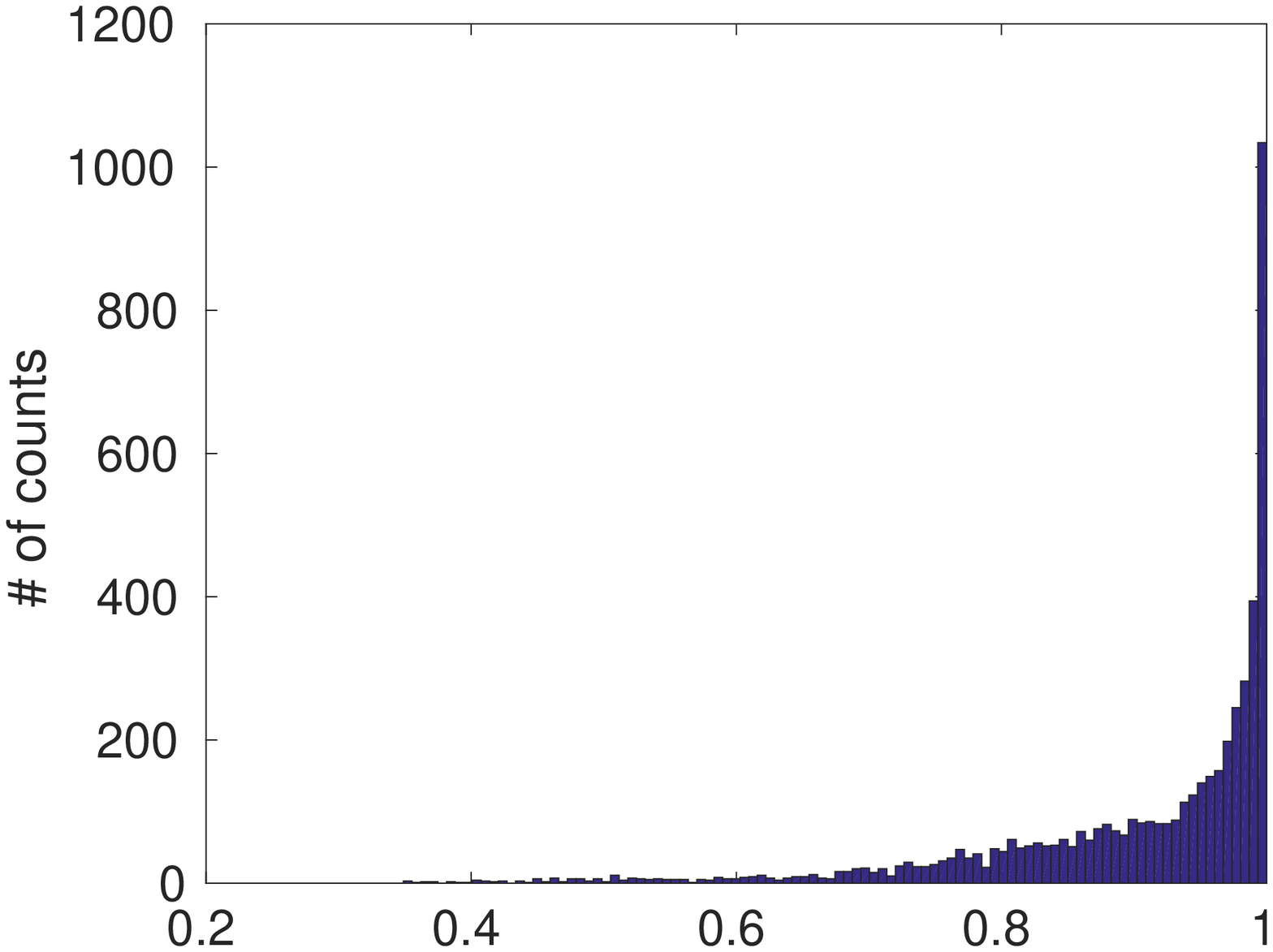}
        \caption{Histogram of $\iota \left( D_2[:,m], \hat{D}_2 \right)\; \forall m$ for test case $A$.}
        \label{fig:block_hist_a_showing_agnostic_effect}
    \end{minipage}
\end{figure}

Next, we evaluate the performance of the MSBDL algorithms in Table \ref{table:taxonomy}. We repeat the bimodal experiment and compare the proposed methods with J$\ell_1$DL, which is the only competing multimodal dictionary learning algorithm that has a stochastic version. The dictionary recovery results are shown in Fig. \ref{fig:stochastic}. The results show that J$\ell_1$DL is not able to recover any part of either the $30$ dB nor $10$ dB dataset dictionaries. In terms of the asymptotic performance as the batch size approaches $L$, MSBDL-1 exhibits negligible bias on both datasets, whereas the other MSBDL flavors incur a small bias, especially on the $10$ dB dataset. On the other hand, it is interesting that MSBDL-2 dramatically outperforms MSBDL-1 for batch sizes less than $300$, which is unexpected since MSBDL-2 performs approximate sufficient statistic computations. The poor performance of MSBDL-3 and MSBDL-4 suggests that these algorithms should be considered only in extremely memory constrained scenarios. Finally, we report on the performance of the proposed annealing strategy for $\sigma_j$. For the bimodal dataset, one expects to $\sigma_1$ to converge to a smaller value than $\sigma_2$. Fig. \ref{fig:hist_s2} shows a histogram of the values to which $\bm{\sigma}$ converge to. The results align with expectations and lend experimental validation for the annealing strategy. We observe the same trend for MSBDL-1 with $L_0 < L$.

\begin{figure}
\centering
\begin{subfigure}{0.45\columnwidth}
\centering
\includegraphics[width=\linewidth]{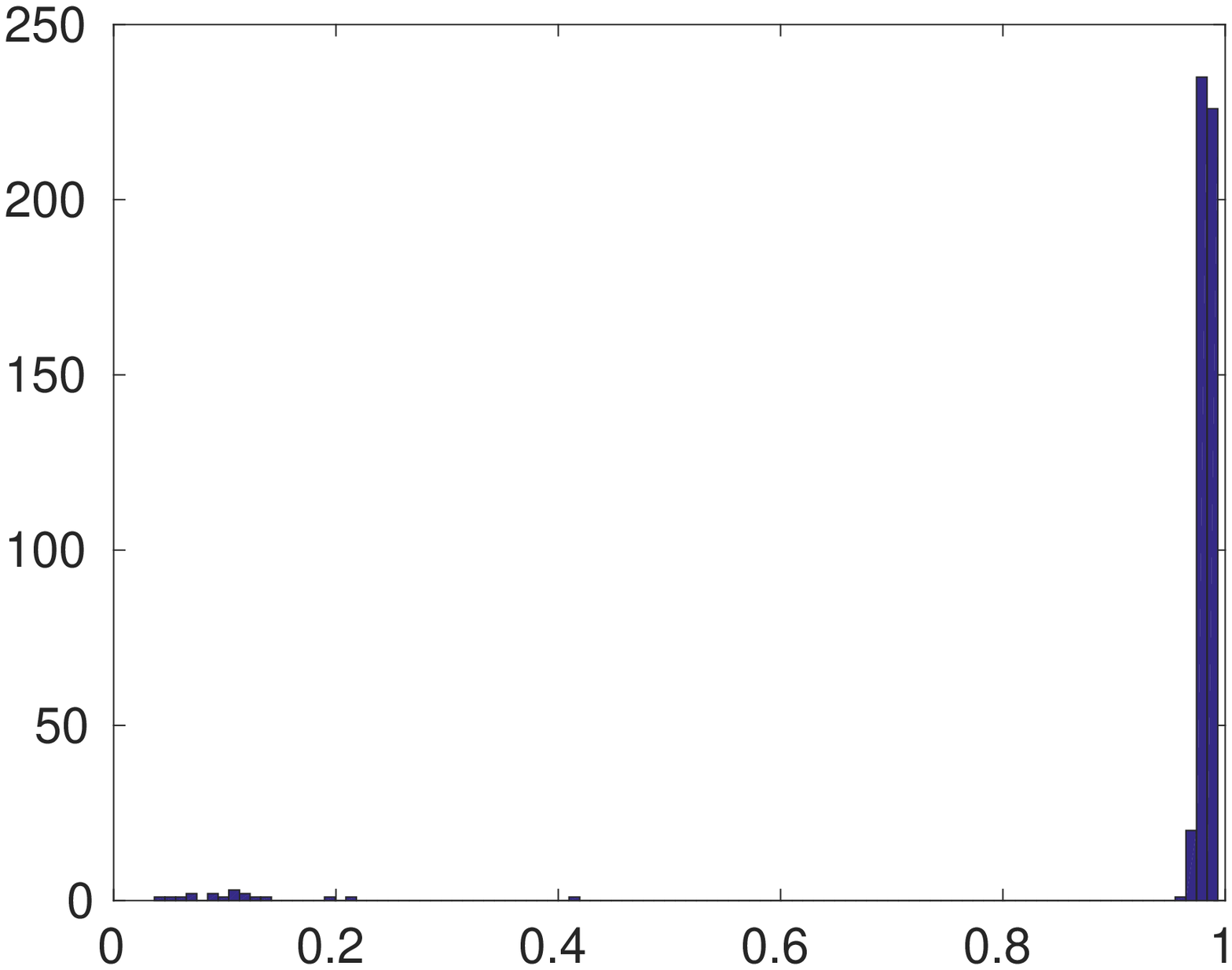}
\caption{}
\label{fig:block_hist_C_2 in main}
\end{subfigure}
~
\begin{subfigure}{0.45\columnwidth}
\centering
\includegraphics[width=\linewidth]{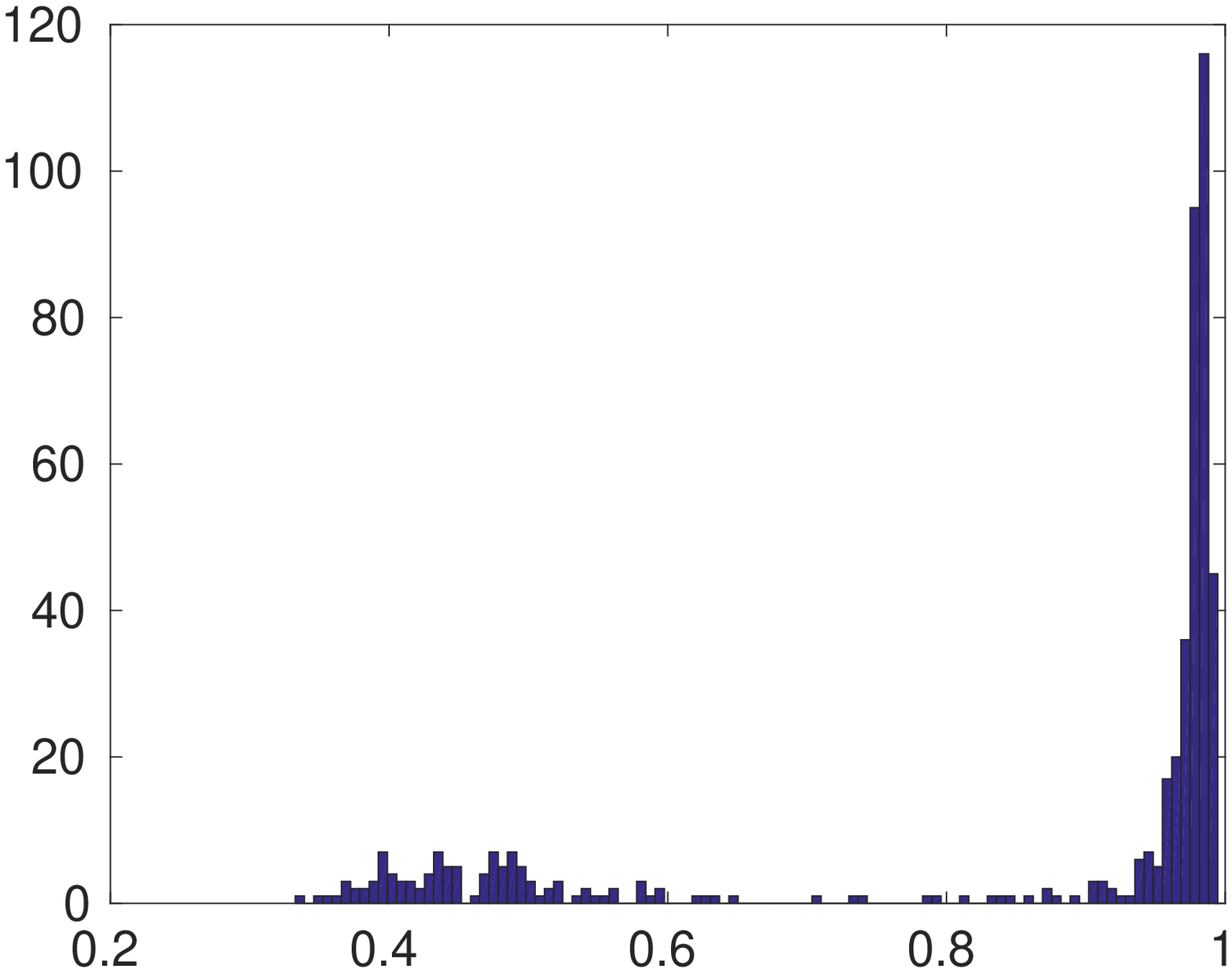}
\caption{}
\label{fig:tree_hist_C_2 in main}
\end{subfigure}
\caption{Histograms of $\function{\iota}{D_2[:,\mathscr{T}^k],\hat{D}_2}$ for test case C in Table \ref{table:block} (\ref{fig:block_hist_C_2 in main}) and $\function{\iota}{D_2[:,m],\hat{D}_2}$ for test case C in Table \ref{table:tree} (\ref{fig:tree_hist_C_2 in main}).}
\end{figure}

To validate the performance of MSBDL using the atom-to-subspace model, we run a number of synthetic data experiments. In all cases, we use $J=2$ and $L = 1000$. We use MSBDL-1 with $L_0 = 500$ to highlight that the algorithm works in incremental EM mode. We simulate $4$ scenarios, summarized in Table \ref{table:block}. For each scenario, we first generate the elements of the ground-truth dictionary $\bm{D}$ by sampling from a $\mathsf{N}\left(0,1 \right)$ distribution and normalizing the resulting dictionaries to have unit $\ell_2$ column norm. We then set $\T^k =
\lbrace k , k+M_1 \rbrace \text{ if } k + M_1 \leq M_2$ and $k \text{ otherwise}$.
This choice of $\braces{\T^k}_{k \in [K]}$ represents the most uniform assignment of columns of $D_2$ to columns of $D_1$. We then generate $x_1^i$ by randomly selecting $s=5$ indices and generating the non-zero entries by drawing samples from a $\mathsf{N}(0,1)$ distribution. We use $\braces{\T^k}_{k \in [K]}$ to find the support of $x_2^i$ and generate the non-zero entries by drawing from a $\mathsf{N}(0,1)$ distribution. In order to assess the performance of the learning algorithm, we must first define the concept of distance between multi-dimensional subspaces. We follow \cite{gunawan2005formula} and compute the distance between $D_2[:,\T^k]$ and $\hat{D}_2$ using 
\begin{align*}
\iota \left( D_2[:,\T^k], \hat{D}_2 \right) = \max_{1 \leq k' \leq K} \sqrt{\vert V_{1}^T V_{2} V_{2}^T V_{1} \vert},
\end{align*}
where we use $D_2[:,\T^k]$ to denote the columns of $D_2$ indexed by $\T^k$, and $V_{1},V_{2}$ denote orthonormal bases for $D_2[:,\T^k]$ and $\hat{D}_2[:,\T^{k'}]$, respectively. We then define the distance between $D_2$ and $\hat{D}_2$ using the two quantities
\begin{align*}
\vartheta_{1}\left(D_2,\hat{D}_2 \right) &= c_{1} \sum_{k \in \lbrace k: \vert \T^k \vert = 1\rbrace} \mathbbm{1}\left[\iota \left( D_2[:,\T^k], \hat{D}_2 \right) > 0.99\right]\\
\vartheta_{2} \left(D_2,\hat{D}_2 \right) &= c_{2} \sum_{k \in \lbrace k: \vert \T^k \vert > 1\rbrace} \mathbbm{1}\left[\iota \left( D_2[:,\T^k], \hat{D}_2 \right) > 0.99\right]
\end{align*}
where $c_{1} =  \vert \lbrace k: \vert \T^k \vert = 1\rbrace  \vert^{-1}$ and $c_{2} = \vert \lbrace k: \vert \T^k \vert > 1\rbrace  \vert^{-1}$. We use MSBDL-1 to learn $\bm{\hat{D}}$, with accuracy results reported in Table \ref{table:block}. Histograms of results are provided in the Supplementary Materials for a higher resolution perspective into the performance of the proposed approach. Note that Table \ref{table:block} reports the accuracy of MSBDL-1 in recovering both the atoms and subspaces of $D_2$. Test case A simulates the scenario where both modalities have a high SNR and $M_2 = 2M_1$. In other words, test case A tests if MSBDL-1 is able to learn in the atom-to-subspace model, without the complications that arise from added noise. The results show that MSBDL-1 effectively learns both the atoms of $D_1$ and the subspaces comprising $D_2$. Test case B simulates the scenario where $\vert \T^k \vert = 1$ for some $k$ but not for others. In effect, test case B tests the pruning strategy described in Section \ref{sec:poor stationary points} and summarized in Fig. \ref{alg:pruning}. The results show that the pruning strategy is effective and allows MSBDL-1 to learn both the atoms of $D_1$ and the atoms and subspaces comprising $D_2$. Test case C is identical to test case B, but with noise added to modality $2$. The results show that MSBDL-1 still effectively recovers $D_1$, but there is a drop in performance with respect to recovering the atoms of $D_2$ and a significant drop in recovering the subspaces of $D_2$. The histogram in Fig. \ref{fig:block_hist_C_2 in main} shows that the distribution of $\braces{\iota(D_2[:\mathscr{T}^k],\hat{D}_2)}_{k \in K}$ is concentrated near $1$ for test case C, suggesting that an alignment threshold of $0.99$ is simply too strict in this case. Finally, test case D demonstrates that MSBDL-1 exhibits robust performance when the modality comprising the roots of the tree is noisy. To provide experimental evidence for the fact that the atom-to-subspace model is agnostic to the atoms of $D_2$, as discussed in Section \ref{sec:subspace2subspace}, we show the ability of MSBDL-1 to recover the atoms of $D_2$ for test case $A$ in Fig. \ref{fig:block_hist_a_showing_agnostic_effect}. The results show that MSBDL-1 is not able to recover the atoms of $D_2$.

Next, we present experimental results for dictionary learning under the hierarchical model, using the same setup as for the atom-to-subspace model. We simulate $4$ scenarios, summarized in Table \ref{table:tree}. To evaluate the performance of the proposed approach, we measure how well it is able to recover the atoms of $D_1$ and $D_2$, where we distinguish between the atoms of $D_2$ corresponding to $\vert \T^k \vert = 1$ and $\vert \T^k \vert > 1$ using

\begin{small}
\begin{align*}
\varrho_{1}\left(D_2,\hat{D}_2 \right) &= c_3 \sum_{k \in \lbrace k: \vert \T^k \vert = 1\rbrace} \mathbbm{1}\left[\iota \left( D_2[:,\T^k], \hat{D}_2 \right) > 0.99\right]\\
\varrho_{2}\left(D_2,\hat{D}_2 \right) &= c_4 \sum_{k \in \lbrace k: \vert \T^k \vert = 1\rbrace, m \in \T^k} \mathbbm{1}\left[\iota \left( D_2[:,m], \hat{D}_2 \right) > 0.99\right]
\end{align*}
\end{small}%
where $c_3 = \vert \lbrace k: \vert \T^k \vert = 1\rbrace  \vert^{-1}$ and $c_4 = \vert \lbrace k: \vert \T^k \vert > 1\rbrace  \vert^{-1}$. The recovery results are reported in Table \ref{table:tree}. Histograms of the results are provided in the Supplementary Material. Test case A demonstrates that MSBDL-1 is able to learn the atoms of $D_1$ and $D_2$ in the low noise scenario for $M_2 = 2M_1$. Test case B shows that MSBDL-1 is able to learn the atoms of $D_1$ and $D_2$ for $M_1 < M_2 < 2M_1$, i.e. highlighting that the pruning strategy in Fig. \ref{alg:pruning} is effective for the hierarchical sparsity model. Test case C adds a considerable amount of noise to the modality occupying the leaves of the tree. Although the recovery results in Table \ref{table:tree} suggest that MSBDL-1 does not perform well in recovering $D_2$ in this scenario, the histogram in Fig. \ref{fig:tree_hist_C_2 in main} shows robust performance. Finally, test case D shows the scenario where a large amount of noise is added to the modality occupying the roots of each subtree.

\begin{table}
\centering
\resizebox{\columnwidth}{!}{%
\begin{tabular}{|cccccccc|}
\hline
& $d_1,M_1$ & $d_2,M_2$ & $SNR_1$ & $SNR_2$ & $\vartheta_{1}(D_1,\hat{D}_1)$ & $\vartheta_{1}(D_2,\hat{D}_2)$ & $\vartheta_{2}(D_2,\hat{D}_2)$\\ \hline
A & $20,50$ & $40,100$ & $30$ & $30$ & $99.8$ & --- & $92$ \\
B & $20,50$ & $30,60$ & $30$ & $30$ & $99.5$ & $100$ & $97$\\
C & $20,50$ & $30,60$ & $30$ & $10$ & $99.9$ & $68.05$ & $3.6$\\
D & $20,50$ & $30,60$ & $10$ & $30$ & $73.64$ & $97.7$  &  $75.4$ \\ \hline
\end{tabular}}
\caption{Recovery results using atom-to-subspace model.}
\label{table:block}
\end{table}

\begin{table}
\centering
\resizebox{\columnwidth}{!}{%
\begin{tabular}{|cccccccc|}
\hline
& $d_1,M_1$ & $d_2,M_2$ & $SNR_1$ & $SNR_2$ & $\varrho_{1}\left( D_1,\hat{D}_1\right)$ & $\varrho_{1}\left(D_2,\hat{D}_2 \right)$ & $\varrho_{2}\left(D_2,\hat{D}_2 \right)$\\ \hline
A & $20,50$ & $40,100$ & $30$ & $30$ & $99.3$ & ---& $96.6$  \\
B & $20,50$ & $30,60$ & $30$ & $30$ & $100$ & $94.7$ & $82.7	$ \\
C & $20,50$ & $30,60$ & $30$ & $10$ & $100$ & $22.1$ & $3.3$ \\
D & $20,50$ & $30,60$ & $10$ & $30$ & $5.2$ & $93.1$ & $61.1$ \\ \hline
\end{tabular}}
\caption{Recovery results using hierarchical model.}
\label{table:tree}
\end{table}

\subsection{Photo Tweet Dataset Classification}
\label{sec:phototweet}
We validate the performance of TD-MSBDL on the Photo Tweet dataset \cite{borth2013large}. The Photo Tweet dataset consists of 603 tweets covering 21 topics. Each tweet contains text and an image with an associated binary label indicating its sentiment. The dataset consists of 5 partitions. We use these partitions to perform 5 rounds of leave-one-out cross-validation, where, during each round, we use one partition as the test set, one as the validation set, and the remaining partitions as the training set. For each round, we process the images by first extracting a bag of SURF features \cite{bay2006surf} from the training set using the MATLAB computer vision system toolbox. We then encode the training, validation, and test sets using the learned bag of features, yielding a 500-dimensional representation for each image \cite{csurka2004visual}. Finally, we compute the mean of each dimension across the training set, center the training, validation, and test sets using the computed means, and perform 10 component PCA to generate a 10-dimensional image representation. We process the text data by first building a 2688 dimensional bag of words from the training set using the scikit-learn Python library. We then encode the training, validation, and test sets using the learned bag of words, normalizing the resulting representations by the number of words in the tweet. We then center the data and perform 10 component PCA to yield a 10-dimensional text representation.

We run TD-MSBDL using incremental EM and approximate posterior sufficient statistic computations, referring to the resulting algorithm as TD-MSBDL-2 in accordance with the taxonomy in Table \ref{table:taxonomy}. Our convention is to refer to the text and image data as modalities $1$ and $2$ respectively. We use $M_j = 40 \forall j$, $L_0 = 200$, $\sigma_{1}^0 = \sqrt{0.01}$, $\sigma_{2}^0 = \sqrt{0.2}$, $\beta_{j}^0 = \sqrt{100} \forall j$, $\sigma^\infty = \sqrt{1e-4}, \alpha_\sigma = \sqrt{0.995}$, $\beta^\infty = \sqrt{1e-2}, \alpha_\beta = \sqrt{0.995}$, and $T^V = 500$. We compare TD-MSBDL with several unimodal and multimodal approaches. We use TD-$\ell_1$DL and D-KSVD to learn classifiers for images and text using unimodal data. For TD-$\ell_1$DL we use the validation set to optimize for $\lambda$ over the set $\lbrace 1e-3,1e-2,1e-1,1 \rbrace$. We also compare with TD-J$\ell_1$DL trained on the multimodal data, using the the same validation approach as for TD-$\ell_1$DL. In all cases, we run training for a maximum of $15e3$ iterations. 

The classification results are shown in Table \ref{table:classification}. Comparing TD-MSBDL with the unimodal methods (i.e. TD-$\ell_1$DL and D-KSVD), the results show that TD-MSBDL achieves higher performance for both feature types. Moreover, it is interesting that TD-J$\ell_1$DL performs worse than TD-$\ell_1$DL, suggesting that it is not capable of capturing the multimodal relationships which TD-MSBDL benefits from.

\begin{table}
	\centering
	\resizebox{\columnwidth}{!}{%
	\begin{tabular}{|ccccc|}
	\hline
	 Feature Type & TD-MSBDL-2 & TD-J$\ell_1$DL & TD-$\ell_1$DL &	D-KSVD  \\ \hline
     Images & \textbf{65.6} & 59.2 & 61.1 & 63.9 \\ 
	Text & \textbf{76} & 73.7	   & 74.1 & 69.4	 \\ \hline
	\end{tabular}}
	
	\caption{Photo tweet dataset classification accuracy $(\%)$.}
	\label{table:classification}
\end{table}

\begin{table}
\centering
\resizebox{\columnwidth}{!}{%
\begin{tabular}{|cccc|}
\hline
	& TD-MSBDL-2 & TD-MSBDL-1 &	TD-MSBDL-1  \\
Prior & One-to-one \eqref{eq:gsm} & Hierarchical \eqref{eq:tree prior} & Atom-to-subspace \eqref{eq:prior block} \\
$d_{1} \times M_{1} / d_{2} \times M_{2}$ & $10 \times 40 / 10 \times 40$ & $10 \times 40 / 20 \times 80$ & $10 \times 40 / 20 \times 80$ \\
Images & 65.6 & 69.2 & 69.5 \\ 
Text & 76 & 74.6 & 74.3\\ 
\hline
\end{tabular}}
\caption{Photo tweet dataset classification accuracy $(\%)$ using TD learning and priors from Section \ref{sec:modeling complex relationships}. Our convention is to designate text as modality 1 and images as modality 2.}
\label{table:classification hierarchical}
\end{table}

%\begin{table}
%	\centering
%	\resizebox{\columnwidth}{!}{%
%	\begin{tabular}{|ccccccc|}
%	\hline
%	  Sparse Coding & SNR & MSBDL-2 & J$\ell_1$DL & $\ell_1$DL & J$\ell_0$DL & K-SVD  \\ \hline
%     Unimodal & 20 & \textbf{16.43} & 20.29  & 20.45 & & 20.09 \\ 
%	 & 10 & \textbf{12.71} & 10.41	   & 10.37 & & 10.01	 \\ 
%	 Multimodal & 20 & & 20.41 & 20.77 & & 18.54\\
%	 & 10 & & 10.45 & 10.50 & & 10.49\\ 
%	 \hline
%	\end{tabular}}
%\end{table}

%\begin{table}
%	\centering
%	\resizebox{\columnwidth}{!}{%
%	\begin{tabular}{|ccccccc|}
%	\hline
%	  Sparse Coding & SNR & MSBDL-2 & J$\ell_1$DL & $\ell_1$DL & J$\ell_0$DL & K-SVD  \\ \hline
%     Unimodal & 20 &  & 20.30  & 20.42 & & 20.07 \\ 
%	 & 10 &  & 10.46 & 10.33 & & 	10.01 \\ 
%	 Multimodal & 20 & & 20.41 & 20.80 & & 18.49\\
%	 & 10 & & 10.49 & 10.46 & & 10.50\\ 
%	 \hline
%	\end{tabular}}
%\end{table}

Finally, we show the efficacy of the priors in Section \ref{sec:modeling complex relationships} in classifying the Photo Tweet dataset. The goal is to show that, by allowing the number of atoms of the image and text dictionaries to be different, the atom-to-subspace and hierarchical sparsity priors lead to superior classification performance. We begin by extracting features from the text and image data as before, with the exception that we use $20$ PCA components to represent images. We then set $M_1$, the text data dictionary size, to $40$ and $M_2$, the image dictionary size, to $80$, corresponding to an oversampling factor $M_j / N_j$ of $4$ for both modalities. We run the TD-MSBDL algorithm with the atom-to-subspace and hierarchical sparsity priors. For the atom-to-subspace prior, the only required modification is to change the update rule of $\gamma^i$ to \eqref{eq:gamma update block}\footnote{We also found it necessary to introduce a post-processing step to the output of the TD-MSBDL algorithm with the atom-to-subspace prior. We output the $W_j^t$ which corresponds to the maximum measured classification accuracy on the validation set during training.}. For the hierarchical sparsity prior, the $\gamma^i$, $D_j$, and $W_j$ update rules are modified to \eqref{eq:gamma update hierarchical}, \eqref{eq:D update hierarchical}, and $\left(W_j\right)^{t+1} = H \transpose{U_j^{TD}} S_j \inv{ S_j^T \left( U_j^{TD} \transpose{U_j^{TD}} + \sum_{i \in [L]} \Sigma_{\hat{x},j}^{TD,i} \right) S_j}$,
respectively.
%We set the algorithm hyperparameters to the same values as in Section \ref{sec:phototweet}, with the exception that $\beta_j^0 = 10 \forall j$ for the atom-to-subspace prior. 
Because there are significant dependencies among the elements in $x_j$ a-priori, we use exact sufficient statistic computation, referring to the resulting algorithm as TD-MSBDL-1. 
The classification results are presented in Table \ref{table:classification hierarchical}, where the TD-MSBDL-2 results with one-to-one prior from Section \ref{sec:phototweet} are shown for reference. The results show a significant improvement in image classification. Although text classification deteriorates slightly, the text classification rate for both atom-to-subspace and hierarchical priors is still higher than the competing methods in Table \ref{table:classification}. Note that this type of experiment, where a different number of dictionary atoms is used for the image and text modalities, is not possible for any of the other dictionary learning approaches considered in this work. 

\subsection{CIFAR10}
The CIFAR10 dataset consists of $32 \times 32$ images from 10 classes \cite{krizhevsky2009learning}. In this experiment, we extract $7 \times 7$ patches from images in the dataset to form $Y_1$. We then form $Y_2$ by adding noise to $Y_1$, i.e. $y_2 = y_1 + \xi, \xi \sim \mathsf{N}\left(0,\sigma_\xi^2 \mathsf{I}\right)$. Our goal is to learn representations $\jbraces{D_j}$ for $\jbraces{Y_j}$, where $J=2$, and to evaluate the quality of those representations. Since ground truth dictionaries do not exist in this scenario, we evaluate the learned $\jbraces{D_j}$ by measuring how well they can denoise test data. 

We begin by learning $\jbraces{D_j}$ from training data. Let $\jbraces{X_j}$ denote the sparse codes for training data $\jbraces{Y}$ under $\jbraces{D_j}$ and let $P$ denote the linear mapping from $X_2$ to $X_1$ satisfying
\begin{align*}
    \argmin_{P} \norm{X_1 - P X_2}_2.
\end{align*}
Note that $X_1 = X_2$ for $\ell_1$DL and K-SVD, such that $P = \mathsf{I}$ for these methods. Given a noisy test point $y_2^{test}$, we first compute the sparse code $x_2^{test}$ using $D_2$, we then form $\hat{x}_1^{test} = P x_2^{test}$, and, finally, we form the denoised image patch using $\hat{y}_1^{test} = D_1 \hat{x}_1^{test}$.

We drew $2000$ patches from one of the classes in the CIFAR10 dataset and corrupted them with $10$dB SNR noise to create the training set $\jbraces{Y_j}$. We then selected $10$ validation and $10$ test images from the same class\footnote{The training, validation, and test sets were all mutually exclusive.}, extracted overlapping $7 \times 7$ patches from each image to form $Y_1^{val}$ and $Y_1^{test}$, and added $10$ dB SNR noise to generate $Y_2^{val}$ and $Y_2^{test}$. We set $M_j = 196 \; \forall j$ and learn $\jbraces{D_j}$ using each of the competing algorithms, performing cross-validation on $\lambda$ for $\ell_1$DL and J$\ell_1$DL and on $\jbraces{\lambda_j}$ for J$\ell_0$DL to maximize the output SNR of the validation set patches $\hat{Y}_1^{val}$. We set $L_0 = 400$, $\sigma_1^0 = 1$, $\sigma_2^0 = \sqrt{2}$, $\sigma^\infty = \sqrt{1e-4}$, $\alpha_\sigma = \sqrt{0.999}$ for MSBDL. During the denoising stage for MSBDL, we form $\jbraces{X_j}$ by solving \eqref{eq:jl1dl} with $\jbraces{D_j}$ fixed to the dictionaries learned by MSBDL and cross-validate over $\lambda$. We emphasize that hyperparameter tuning is not necessary to learn $\jbraces{D_j}$ using MSBDL, but is useful during the denoising stage because the dictionary learning task is not necessarily aligned with the denoising task. In particular, since cross-validation was only necessary after $\jbraces{D_j}$ were learned, the learning complexity analysis in Section \ref{sec:learning complexity}, visualized in Fig. \ref{fig:learning complexity figure}, still applies to this experiment. 

Denoising performance on the test set is shown in Table \ref{table:cifar denoising}. MSBDL performed best across all algorithms tested. These results indicate that MSBDL was able to leverage information from both modalities to learn a better representation for $Y_2$ compared to the competing algorithms. On the whole, multimodal learning algorithms outperformed the unimodal algorithms on this task. 

\begin{table}
	\centering
	
	\begin{tabular}{|ccccc|}
	\hline
	 MSBDL-1 & J$\ell_1$DL & J$\ell_0$DL& $\ell_1$DL &	K-SVD  \\ \hline
     \textbf{15.97}  & 15.55 & 15.83 & 15.45 & 12.99 \\ \hline
	\end{tabular}
	
	\caption{CIFAR10 denoising results. The input SNR is 10 dB and the results denote the output SNR in dB.}
	\label{table:cifar denoising}
\end{table}

% \subsubsection{Classification}
% \begin{table}
% 	\centering
% 	\resizebox{\columnwidth}{!}{%
% 	\begin{tabular}{|ccccc|}
% 	\hline
% 	 SNR (dB)  & TD-MSBDL-2 & TD-J$\ell_1$DL & TD-$\ell_1$DL &	D-KSVD  \\ \hline
%      30 & \textbf{93.14}  & 92.77& 93.09 & 92.88\\ 
%      25 & 91.87 & 92.19 & 92.40 & \textbf{92.30}\\
%      20 & {93.19} & 93.14 & \textbf{93.30} & 92.56\\
%      15 & 93.35 & \textbf{93.98} & 93.61 & 93.83\\
%      10 & {93.67} & 93.03 & \textbf{93.83} & 92.93\\ \hline
% %     All & 99.95 & \textbf{100} & &\\ \hline
% 	\end{tabular}}
	
% 	\caption{CIFAR10 classification results.}
% 	\label{table:classification}
% \end{table}

% \begin{table}
% 	\centering
% 	\resizebox{\columnwidth}{!}{%
% 	\begin{tabular}{|ccccc|}
% 	\hline
% 	 SNR (dB)  & TD-MSBDL-2 & TD-J$\ell_1$DL & TD-$\ell_1$DL &	D-KSVD  \\ \hline
%      30 &   &  92.84& &\\ 
%      25 &  & 91.26 & & \\
%      20 &  & 93.39 & & \\
%      15 &  & 93.80 & & \\
%      10 &  & 93.50 & & \\
%      All &  & 100 & &\\ \hline
% 	\end{tabular}}
	
% 	\caption{CIFAR10, d = 20, M = 60}
% 	\label{table:classification}
% \end{table}

% AccTest =
% 0.9247
% 0.9221
% 0.9285
% 0.9381
% 0.9188

% 0.9333
% 0.9247
% 0.9301
% 0.9375
% 0.9178

% mm: 0.9957

% jl1dl: 0.9290    0.9204    0.9327    0.9349    0.9237 1.0

\subsection{MIR Flickr Image Annotation}
\label{sec:mirflickr}
The MIR Flickr dataset consists of images and associated annotations, with each image annotated by up to 38 tags \cite{everingham2007pascal}. For this task, we use 3 kinds of features to represent a given image: 100-dimensional dense hue, 1000-dimensional Harris SIFT, and 4096-dimensional HSV, comprising modalities one, two, and three, respectively \cite{guillaumin2010multimodal}. We use PCA to reduce the feature dimensionality to $N_1 = 20$, $N_2 = 40$, and $N_3 = 60$ for modalities one through three. The choice of $\jbraces{N_j}$ serves to reduce the data dimensionality while preserving the relative cardinality of the modalities, i.e. $N_1 > N_2 > N_3$. Our train and validation sets both contain 2000 images, while the test set contains 12500 images. 

Since each image is annotated by multiple tags, each label vector $h^i$ contains a $1$ for all of the tags associated with the $i$'th image and we denote the total number of tags for image $i$ as $k^i$. In order to evaluate how well each method is able to predict tags from images, we compute $\hat{h}^i = W_j x_j^i$, set all of the elements except the top $k^i$ to $0$, and set the top $k^i$ elements to $1$. We then compute the recall for image $i$ using $\norm{h^i - \hat{h}^i}_1 / \norm{h^i}_1$, which essentially measures the fraction of the true tags for image $i$ appearing in the top $k^i$ elements of $W_j x_j^i$.

We set the dictionary size to $M = 4 N_1$ for TD-J$\ell_1$DL, TD-$\ell_1$DL, and D-KSVD. We learn dictionaries for each modality in a multimodal fashion using TD-MSBDL-1 and TD-J$\ell_1$DL and in a unimodal fashion using TD-$\ell_1$DL and D-KSVD, identical to the setup in Section \ref{sec:phototweet}. Since TD-MSBDL can assign different numbers of dictionary elements to different modalities, we choose the atom-to-subspace prior detailed in Section \ref{sec:subspace2subspace} and set $M_1 = 4 N_1$ and $M_3 = M_2 = 4 N_2$, where each atom of $D_1$ is grouped with two atoms from $D_2$ and two atoms from $D_3$. TD-MSBDL is the only algorithm capable of modeling dictionaries with varying size across modalities, which is useful in this application given that the feature size varies drastically across modalities. We perform cross-validation to set $\lambda$ for TD-J$\ell_1$DL and TD-$\ell_1$DL using the same settings as in Section \ref{sec:phototweet}. We set $s=10$ for D-KSVD and run all algorithms for a maximum of $15e3$ iterations with a batch size of $400$. We set $\sigma_1^0 = \sigma_2^0 = \sigma_3^0 = \sqrt{0.1}, \alpha_\sigma = \sqrt{0.999}, \beta_1^0 = \beta_2^0 = \beta_3^0 = \sqrt{100}, \beta^\infty = \sqrt{1e-4}, \alpha_\beta = \sqrt{0.999}$ for MSBDL.

Table \ref{table:mirflickr} reports the average recall for all of the images in the test set across the algorithms tested. TD-MSBDL performs best across all feature types among the algorithms tested. In particular, TD-MSBDL outperforms the unimodal learning algorithms whereas TD-J$\ell_1$DL performs worse than the unimodal algorithms for the Harris SIFT and HSV feature types.

% \begin{table}
% 	\centering
% 	\resizebox{\columnwidth}{!}{%
% 	\begin{tabular}{|ccccc|}
% 	\hline
% 	 Feature  & TD-MSBDL-1 & TD-J$\ell_1$DL & TD-$\ell_1$DL &	D-KSVD  \\ \hline
%      Dense hue & {0.36}  & 0.41&  &  0.28\\ 
%      Harris sift & {0.41} & 0.37 &  & 0.38\\
%      HSV & {0.38} & 0.39 & & {0.35}\\ \hline
% 	\end{tabular}}
	
% 	\caption{Average recall on MIRFLICKR dataset.}
% 	\label{table:mirflickr}
% \end{table}

% \begin{table}
% 	\centering
% 	\resizebox{\columnwidth}{!}{%
% 	\begin{tabular}{|ccccc|}
% 	\hline
% 	 Feature  & TD-MSBDL-1 & TD-J$\ell_1$DL & TD-$\ell_1$DL &	D-KSVD  \\ \hline
%      Dense hue & {0.39}  & 0.41&  &  0.28\\ 
%      Harris sift & {0.40} & 0.46 &  & 0.38\\
%      HSV & {0.42} & 0.43 & & {0.35}\\ \hline
% 	\end{tabular}}
	
% 	\caption{Average recall on MIRFLICKR dataset.}
% 	\label{table:mirflickr}
% \end{table}

\begin{table}
	\centering
	\resizebox{\columnwidth}{!}{%
	\begin{tabular}{|ccccc|}
	\hline
	 Feature  & TD-MSBDL-1 & TD-J$\ell_1$DL & TD-$\ell_1$DL &	D-KSVD  \\ \hline
     Dense hue & \textbf{0.41}  & \textbf{0.41}& 0.38  &  0.29\\ 
     Harris SIFT & \textbf{0.45} & 0.37 & 0.38  & 0.40\\
     HSV & \textbf{0.41} & 0.39 & \textbf{0.41} & {0.40}\\ \hline
	\end{tabular}}
	
	\caption{Average recall on MIR Flickr dataset.}
	\label{table:mirflickr}
\end{table}

\subsection{Discussion}
\label{sec:discussion}
The results in this section have shown that MSBDL is able to outperform competing dictionary learning methods on a number of tasks, including dictionary recovery, sentiment classification, image denoising, and image annotation. We conclude the results section by reiterating the main points of distinction between MSBDL and competing approaches, which may not be evident by inspecting performance metrics alone.

For all of the experiments performed in this section, MSBDL performed automatic hyperparameter tuning, whereas hyperparameters were tuned using a grid search for all of the competing algorithms\footnote{The exception to this was when we performed cross-validation for the CIFAR10 denoising experiment, but we only did so after having learned $\jbraces{D_j}$, such that the argument in this section still applies.}. This highlights the fact that the proposed approach offers superior performance at a discount, so to speak, in terms of LC. The same LC benefits apply for TD-MSBDL, i.e. the LC of TD-MSBDL is identical to that of MSBDL.

None of the competing methods, including the MSBDL variant in \cite{fedorov2017mutlimodal}, are capable of learning models where the number of dictionary elements varies across modalities. This was particularly useful in the experiments on the Photo Tweet and MIR Flickr datasets, where the data dimensionality varied significantly across modalities. 

\section{Conclusion}
We have detailed a sparse multimodal dictionary learning algorithm. Our approach incorporates the main features of existing methods, which establish a  correspondence between the elements of the dictionaries for each modality, while addressing the major drawbacks of previous algorithms. Our method enjoys the theoretical guarantees and superior performance associated with the sparse Bayesian learning framework.

\bibliographystyle{IEEEtran}
%\bibliography{ManuscriptBib}

\end{document}